\documentclass[journal=jacsat,manuscript=article]{achemso}

\usepackage[version=3]{mhchem}

\usepackage{geometry}
\usepackage[utf8]{inputenc}
\usepackage[T1]{fontenc}
\usepackage{natbib}
\usepackage{amsmath}
\usepackage{amsfonts}
\usepackage{microtype}
\usepackage{graphicx}
\usepackage{subfigure}
\usepackage{booktabs} 
\usepackage{url} 
\usepackage[colorlinks=True]{hyperref}
\usepackage{multirow}
\usepackage{times}
\usepackage{mathtools}
\usepackage{textcomp}
\usepackage{booktabs}
\usepackage[normalem]{ulem}
\usepackage{tabularx}
\usepackage{makecell}
\usepackage{balance}
\usepackage{xcolor}
\usepackage{float}
\usepackage{algorithm}
\usepackage{algorithmic}
\usepackage{soul}
\usepackage{xparse,soul}
\usepackage{cancel}

\newcolumntype{Y}{>{\centering\arraybackslash}X}

\newcommand\sj[1]{{\color{black}#1}}
\newcommand\ms[1]{{\color{black}#1}}

\makeatletter
\NewDocumentCommand{\sotwo}{O{red}O{black}+m}
    {%
        \begingroup
        \color{#1}%
        \setul{-.5ex}{.4pt}%
        \def\SOUL@uleverysyllable{%
            \rlap{%
                \color{#2}\the\SOUL@syllable
                \SOUL@setkern\SOUL@charkern}%
            \SOUL@ulunderline{%
                \phantom{\the\SOUL@syllable}}%
        }%
        \ul{#3}%
        \endgroup
    }

\definecolor{darkblue}{rgb}{0.0,0.0,0.55}
\hypersetup{
  colorlinks = true,
  citecolor  = darkblue,
  linkcolor  = darkblue,
  citecolor  = darkblue,
  filecolor  = darkblue,
  urlcolor   = darkblue,
}
\makeatletter
\let\@fnsymbol\@arabic
\makeatother
\date{}

\title{Molecule Edit Graph Attention Network: \\
           Modeling Chemical Reactions as Sequences of Graph Edits}
\author{Mikołaj Sacha}
\affiliation[Molecule One]
{Molecule One, Poland}
\author{Mikołaj Błaż}
\affiliation[Molecule One]
{Molecule One, Poland}
\author{Piotr Byrski}
\affiliation[Molecule One]
{Molecule One, Poland}
\author{Paweł Dąbrowski-Tumański}
\affiliation[Molecule One]
{Molecule One, Poland}
\alsoaffiliation[UKSW]
{Faculty of Mathematics and Natural Sciences. School of Exact Sciences, Cardinal Stefan Wyszynski University, Poland}
\author{Mikołaj Chromiński}
\affiliation[UW]
{Centre of New Technologies, University of Warsaw, Poland}
\author{Rafał Loska}
\affiliation[IPAN]
{Institute of Organic Chemistry, Polish Academy of Sciences, Poland}
\author{Paweł Włodarczyk-Pruszyński}
\affiliation[Molecule One]
{Molecule One, Poland}
\author{Stanisław Jastrzębski}
\affiliation[Molecule One]
{Molecule One, Poland}
\alsoaffiliation[Jagiellonian University]
{Jagiellonian University, Poland}
\alsoaffiliation[New York University]
{New York University, USA}
\email{stan@molecule.one}

\begin{document}

\maketitle

\begin{abstract}

The central challenge in automated synthesis planning is to be able to generate and predict outcomes of a diverse set of chemical reactions. In particular, in many cases, the most likely synthesis pathway cannot be applied due to additional constraints, which requires proposing alternative chemical reactions. With this in mind, we present Molecule Edit Graph Attention Network (MEGAN), an end-to-end encoder-decoder neural model. MEGAN is inspired by models that express a chemical reaction as a sequence of graph edits, akin to the arrow pushing formalism. We extend this model to retrosynthesis prediction (predicting substrates given the product of a chemical reaction) and scale it up to large datasets. We argue that representing the reaction as a sequence of edits enables MEGAN to efficiently explore the space of plausible chemical reactions, maintaining the flexibility of modeling the reaction in an end-to-end fashion, and achieving state-of-the-art accuracy in standard benchmarks. Code and trained models are made available online at \href{https://github.com/molecule-one/megan}{https://github.com/molecule-one/megan}.

\end{abstract}

\section{Introduction}
\label{introduction}

Synthesis planning answers the question of \emph{how} to make a given molecule. Due to the substantial size and complexity of the reaction space, synthesis planning is a demanding task even for skilled chemists and remains an important roadblock in the drug discovery process~\citep{blake2018}. Computer-aided synthesis planning (CASP) methods can speed-up the drug discovery process by assisting chemists in designing syntheses~\citep{corey,segler2018,coley_ml_in_synth,mt_unifies}. These methods are also increasingly employed in \textit{de novo} drug design, in which estimating the synthetic accessibility of a large number of compounds has surfaced as a key challenge to the field~\citep{coley2020}. 

Designing a synthesis plan involves predicting reaction outcomes for a given set of possible substrates (\emph{forward synthesis prediction}), or proposing reactions that can simplify a given target molecule (\emph{retrosynthesis prediction})~\citep{warren2007organic}. In both cases, it is key for automated synthesis planning tools to model reliably outcomes of a diverse set of chemical reactions. This is especially important when there are additional constraints imposed on the desired synthesis plan such as avoiding certain starting materials or using green chemistry~\citep{struble2020}.

Many approaches to predicting reaction outcomes employ a static library of \emph{reaction templates}. A reaction template encodes a graph transformation rule that can be used to generate the reaction based on the product or subtrates~\citep{Corey178,Satoh1995Sophia,segler2017,coley2017,gln,Grzybowski2018ChematicaAS}. Due to computational limitations, they typically require representing reactions using a restricted number of templates, which necessarily limits the \emph{coverage} of the chemical reaction space accessible by such methods. 

The small coverage and other limitations of template-based methods motivated the development of more flexible deep learning models, which aim to learn transformations and their applicability rules directly from the data~\citep{mt, mt_retro}. Despite the added flexibility and the resulting better coverage of the chemical space, they tend to propose a limited set of plausible chemical reactions for a given input~\citep{chen2019learning}. Models such as proposed in \citet{mt, mt_retro} belong to the class of sequence to sequence models. They generate a chemical reaction by sequentially outputting individual symbols in the SMILES notation~\citep{smiles}. We hypothesize this makes it challenging to generate more diverse predictions from the model, as there is no natural decomposition of the predictive distribution into plausible but different reactions. Notably, many previous papers have pointed out limited diversity of sequence to sequence models in other domains~\citep{roberts2020decoding,he-etal-2018-sequence,jiang-de-rijke-2018-sequence}.

An arguably more natural idea is to express a reaction as a sequence of graph edits such as bond addition or removal, which is inspired by how chemists describe reactions using the arrow pushing mechanism~\citep{electron,kien2019}.  \citet{electron} introduce a model that predicts the reaction mechanism represented as a sequence of bond removal and additions that directly correspond to electron paths. However, their approach is limited to a certain subset of the chemical reaction space (reactions with linear chain topology) and forward synthesis. In a similar spirit, \citet{kien2019} models a reaction as a set of operation on atom pairs. However, due to the lack of support for atom addition, their method also cannot be readily applied to retrosynthesis prediction. We hypothesize that expressing reaction in this way enables to easily search through different alternative reactions because the proposed reaction is obvious usually after the first few generation steps. This allows the generation algorithm to naturally backtrack and explore other possible reactions. We note that related methods have a rich history in parallel applications of generative deep learning such as natural language processing~\citep{gui2019}.

In this work, we propose the Molecule Edit Graph Attention Network (MEGAN), an encoder-decoder end-to-end model that generates a reaction as a sequence of graph edits. Designing a novel architecture and training procedure enables us to apply this approach to both retro and forward synthesis, and scale it up to large datasets. More specifically, our two main contributions are as follows:
\begin{itemize}
    \item We extend the idea to model the reaction as a sequence edits to retrosynthesis prediction and scale it to large datasets. This required introducing a novel encoder-decoder architecture and an efficient training procedure that avoids the need to use reinforcement learning.
    \item We achieve competitive performance on retrosynthesis and competitive performance on forward synthesis as well as state-of-the-art top-k accuracy for large $K$ values on all tested datasets. This serves as evidence that MEGAN achieves excellent coverage of the reaction space.
\end{itemize}

We open-sourced our code and trained models at \href{https://github.com/molecule-one/megan}{https://github.com/molecule-one/megan}.

\section{Related work}

Computer-aided reaction prediction has a rich history~\citep{Todd2005}. Early approaches to generating reactions relied on manually crafted rules~\citep{salatin1980}, which were difficult to apply to novel chemistry. Another family of methods uses physical chemistry calculations~\citep{zimmerman2013}. These methods are typically too computationally intensive to be used in synthesis planning software and can be complemented using statistical methods that learn from data.

Statistical approaches to predicting reactions and designing synthesis paths can be broadly categorized into template-based and template-free approaches. Template-based approaches use reaction rules or templates. These templates can be automatically mined from a database of known reactions or defined manually~\citep{Corey178,Satoh1995Sophia,segler2017,coley2017,gln,Grzybowski2018ChematicaAS}. 

Manually encoding reaction templates necessitates developing complex rules governing their applicability~\citep{Grzybowski2018ChematicaAS,grzyb}. This motivated the recent rise in popularity of machine-learning models used to select or rank most relevant templates~\citep{guzik2016,segler2018}. Such models can be applied to assess the reactivity of atoms to which reaction rules should be applied~\citep{weisfeiler, coley2018reactivity}. An interesting alternative to this approach is to prioritize transformations that have already been applied to similar molecules~\citep{retrosim}. A suitable set of templates combined with a strong ranking model can achieve state-of-the-art performance on standard retrosynthesis benchmarks~\cite{gln}.

The small coverage and other limitations of template-based methods motivated the development of template-free models. A particularly successful class of template-free methods is based on sequence to sequence models that sequentially (symbol by symbol) predict the target SMILES~\citep{smiles} string from the input SMILES string~\citep{smiles}. Such approaches can be used out-of-the-box for both forward synthesis~\citep{schwaller_seq2seq, mt} and retrosynthesis~\citep{seq2seq, mt_retro}. However, they have been shown to make some trivial mistakes~\citep{grzyb}, and produce reactions with a limited diversity~\citep{chen2019learning}. These models also act as black-boxes; they do not provide the reasoning behind their predictions and are not able to map atoms between the substrates and the product. 

We hypothesize that the limited diversity of sequence to sequence models can be attributed to the fact that they generate reaction as a sequence of SMILES symbols from left to right. Many papers have also pointed out shortcomings in the diversity of the generated output by sequence to sequence models~\citep{roberts2020decoding,he-etal-2018-sequence,jiang-de-rijke-2018-sequence}. In natural language processing, there is a rich parallel history of exploring alternative generation ordering and adding through various means structure to the output.  \citet{gui2019} show that using an alternative order of output symbols than left to right enables generating more diverse prediction. \citet{Mehri2018MiddleOutD} propose an alternative decoding algorithm to improve the diversity of sequence to sequence models that starts from generating the middle word. These modeling choices can be viewed as incorporating more suitable inductive biases for the given domain.  From this perspective, we study an alternative representation of the output in generative models applied to reaction outcome prediction. 

Our work focuses on a recently introduced class of template-free methods, which defines reaction generation as predicting target graph by sequentially modifying the input graph. Such an approach, while remaining template-free, can provide greater interpretability of predictions, which are modeled as direct transformations on molecules. Models that predict graph edits have been successfully applied to forward synthesis~\citep{kien2019} and can be fine-tuned to provide especially interpretable results on certain subsets of the chemical reaction space~\citep{electron}. In this work, we propose a model inspired by these prior works and extend the approach to retrosynthesis prediction.

Concurrently to our work, the graph-edit reaction generation approach has been employed to retrosynthesis {in \textsc{G2Gs} and \textsc{GraphRetro} models}~\citep{g2g,graph_retro,yan2020}. The {\textsc{G2Gs}} model presented in \citet{g2g,yan2020} consists of two separate modules for predicting the reaction center and generating final substrates from the disconnected synthons. {\textsc{GraphRetro}\citep{graph_retro}} employs a similar framework but completes the synthons with substructures which are selected from a predefined set found on the training data. In contrast to these concurrent works, we propose an end-to-end model for generating reaction through a sequence of edits and apply it both to retro and forward synthesis prediction. 

\sj{Our work on reaction outcome prediction is motivated by their wide utility. An important use of reaction outcome prediction models arises in automated synthesis planning; see for example \citet{segler2018} for an excellent discussion. It is also worth noting that reaction outcome prediction has recently risen in importance in the context of de novo drug discovery. \citet{gao2020} have drawn attention to the fact that many recent de novo drug discovery methods tend to generate hard to synthesize molecules. In this context, automated synthesis planning can be used to filter out these hard to synthesize compounds, and compares favorable to other simpler methods~\citep{liu2020retrognn,Thakkar2021}. Other works use reaction outcome prediction models to define the space of accessible molecules to de novo models~\citep{Gottipati2021}. }

\section{Molecule Edit Graph Attention Network}
\label{sec:megan}
\begin{figure*}
\begin{center}
\includegraphics[width=0.95\textwidth]{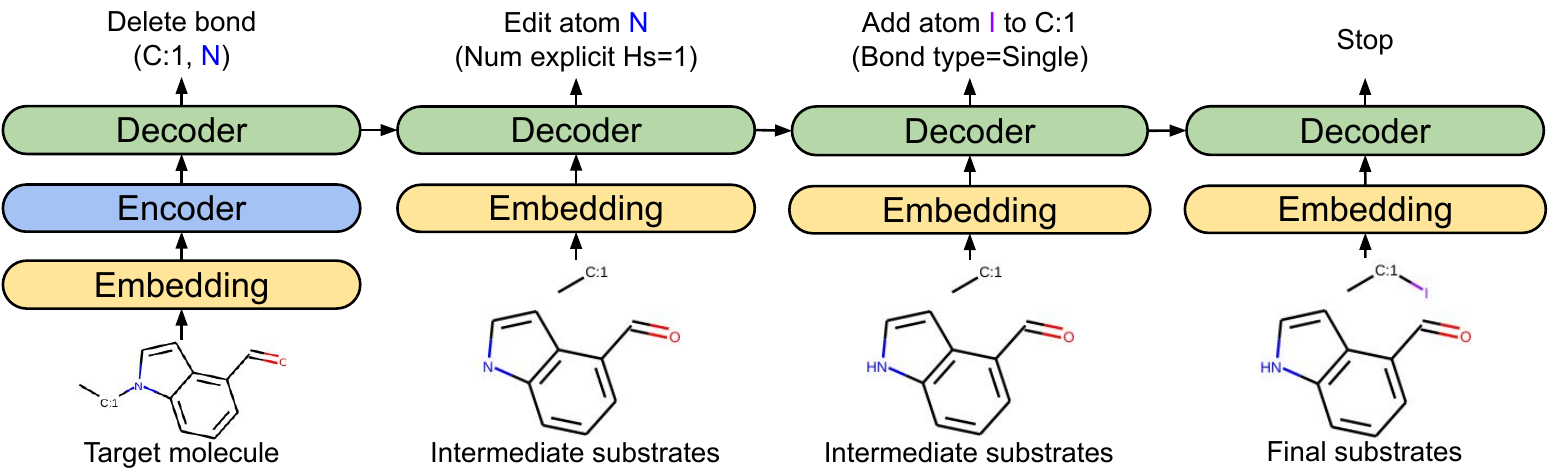}
\caption{\sj{In retrosynthesis prediction task, Molecule Edit Graph Attention Network (MEGAN) generates substrates by sequentially modifying the target molecule. In this example, in the first step the model decides to delete a bond between the carbon and the nitrogen atom. The generation process ends when MEGAN outputs the stop action. MEGAN is an encoder-decoder model, where the encoder and the decoder are constructed by stacking graph convolutional layers based on Graph Attention Network~\citep{gat}. Before passing the graph to the encoder or decoder, we embed the atoms and bonds using one-hot encoded features, which are projected down by two separate linear layers. We describe the remaining details about the model and the used representation in Section~\nameref{sec:input-output-representation} and Section~\nameref{sec:gradientbasedtraining}}}
\label{fig:megan_drawing}
\end{center}
\end{figure*}

Molecule Edit Graph Attention Network (MEGAN) is an encoder-decoder architecture based on \sj{the graph convolutional network architecture} that predicts a sequence of actions on atoms and bonds of a chemical compound. 

\sj{

\subsection{Input and output representation}
\label{sec:input-output-representation}

We begin by describing the input and the output representation of the MEGAN model.

\paragraph{Input representation}
MEGAN takes as input a molecular graph, which is represented by labeled node vectors} \ms{ (atoms)} \sj{ and a labeled adjacency matrix} \ms{ (bonds)} \sj{. 

The input $x = (\mathbf{H}^{OH}, \mathbf{A}^{OH})$ consists of a matrix of one-hot encoded atom features } \ms{$\mathbf{H}^{OH} \in {\mathbb{Z}_{\geq 0}}^{n\times n_a}$} \sj{ and one-hot encoded bond features $\mathbf{A}^{OH} \in {\mathbb{Z}_{\geq 0}}^{n\times n\times n_b}$, where $n$ is the number of nodes in a graph and $n_a$ and $n_b$ are sizes of atom and bond feature vectors. Hydrogen atoms are removed from the graph, as for heavy atoms the number of neighboring hydrogen atoms can be deduced implicitly. 

We add an additional node to the graph named \emph{supernode}~\citep{supernode}, which is connected to all atoms in the graph with a special \textsc{Supernode} bond type. In graph convolutional neural networks, supernode is particularly useful for passing information between connected components of a graph, especially after it was split by deleting a bond. \ms{ Adding the supernode makes $n$ equal to the number of atoms in the input molecule + 1.}

\sj{To construct feature vectors for atoms and bonds , we use the following chemical properties calculated using the RdKit package~\citep{rdkit}: \textsc{atomic number}, \textsc{formal charge}, \textsc{chiral tag}, \textsc{number of explicit hydrogen atoms}, \textsc{is aromatic} for atoms and \textsc{bond type}, \textsc{bond stereometry} (None, Z or E) \ms{ for bonds}. All these features are set to zero for the supernode and all bonds with the supernode, except for \textsc{bond type} which has a special \emph{supernode} value for bonds between atoms and the supernode. We also add a binary flag to indicate whether a given atom or bond was edited or added in any of the previous steps. Finally, we indicate whether a given node is an atom or the supernode with a binary flag on each node. Final atom and bond vectors are constructed by concatenating one-hot vectors of all of their categorical features. We describe the remaining details in Section \nameref{supp:featurization} in the Supplement.}

\paragraph{Output representation}

\sj{The output of MEGAN is a sequence of graph actions, which are then applied to the input graph to produce the desired output graph}. \ms{An action can be applied to a single atom or a pair of atoms. In the latter case the action modifies the bond between the two atoms. } To train the model, we compare the outputted sequence of actions with the ground-truth sequence based on a mapping between the product and the substrates (see Section~\nameref{sec:gradientbasedtraining}). We define the following graph actions:

\begin{itemize}
\item Edit atom properties \emph{(EditAtom)}
\item Edit bond between two atoms \emph{(EditBond)}
\item Add new atom to the graph \emph{(AddAtom)}
\item Add new benzene ring to the graph \emph{(AddBenzene)}
\item Stop generation \emph{(Stop)}
\end{itemize}

\emph{EditAtom} changes properties of atoms, such as the formal charge, chirality, or aromaticity. \emph{EditBond} adds, edits, or deletes a bond between two atoms. \emph{AddAtom} adds a new atom of a specified type as a neighbor of another atom already existing in the graph, with a specified bond type. \emph{AddBenzene} simplifies generation by allowing the model to append a complete benzene ring to a selected carbon atom. \emph{Stop} action indicates the end of the generation process \ms{and can be outputted only by the supernode}. We describe the possible actions in detail in the "Graph edit actions" in Supplement. 
f
\subsection{Model architecture}
\label{sec:model-architecture}

\sj{\paragraph{Encoder-decoder architecture of MEGAN}
The MEGAN model consists of two parts: the \emph{encoder}, which is invoked only once per reaction generation at its beginning, and the \emph{decoder}, which is sequentially invoked to generate actions. This general encoder-decoder architecture is shared by the most successful sequence to sequence models in deep learning~\citet{attention}. 

Both encoder and decoder are variants of graph convolutional networks (see \nameref{par:gcn_att}) that modify hidden vector representation of nodes. They take the same input and output node feature vector size $n_a$ and input edge feature vector size $n_b$. The input graph $x_t$ is modified according to the action predicted by the decoder at each step $t$, until a special \emph{Stop} action is outputted. We stack $n_e$ and $n_d$ layers of the encoder and decoder, respectively. The architecture and the inference process are depicted schematically in Figure~\ref{fig:megan_drawing}.

\paragraph{Generating graph actions}
At each generation step $t$, the one-hot encoded features of the current graph $x_t=(\mathbf{H}^{OH}_t, \mathbf{A}^{OH}_t)$ are first embedded using two linear layers $\mathbf{H}^0_t = f_{emb}(\mathbf{H}^{OH}_t)$ and $\mathbf{A}_t=g_{emb}(\mathbf{A}^{OH}_t)$ to convert atoms and bonds to $n_a$ and $n_b$-dimensional spaces, respectively. We treat $n_a$ and $n_b$ as hyperparameters. 

In the first generation step ($t=1$), we pass the embedded graph through the encoder and the decoder to arrive at atom features $\mathbf{H}_1=\textrm{dec}(\textrm{enc}(\mathbf{H}^0_1))$.  At each subsequent step $t$, we only use the decoder. We incorporate previous step representation by taking the element-wise maximum operation over the current and previous atom features. Namely, } \ms{for $t>1$, } \sj{$\mathbf{H}_t = \textrm{dec}(\max(\textrm{enc}(\mathbf{H}_{t-1}), \mathbf{H}_{t-1}))$. We zero-pad features of $\mathbf{H}_{t-1}$ for any node that was added to the graph at step $t$. The encoder and decoder also take as input $\mathbf{A}_{t}$.

The logits $\mathbf{L}^a_t$ and $\mathbf{L}^b_t$ for atom actions $Act_{a}$ and bond actions $Act_{b}$ are calculated at each step $t$ from $\mathbf{H}_t$ and $\mathbf{A_t}$ as follows:

\begin{align}
    \mathbb{R}^{d} \ni \mathbf{F}_t&= \sigma_r(g_{\mathrm{atom}}(\mathbf{H}_t)) \\
    \mathbb{R}^{|Act_{a}|} \ni \mathbf{L}^a_t &= g'_{\mathrm{atom}}(\mathbf{F}_t) \\
    \mathbb{R}^{d} \ni \mathbf{J}_i &= \sigma_r(g_{\mathrm{bond}}(\mathbf{H}_t)) \\
    \mathbb{R}^{d+n_a} \ni \mathbf{J}'_{ij} &= \sigma_r(\mathbf{J}_i + \mathbf{J}_j \parallel \mathbf{A}_{i, j}) \\
    \mathbb{R}^{|Act_{b}|} \ni \mathbf{L}^b_{tij} &= g'_{\mathrm{bond}}(\mathbf{J}'_{ij})
\end{align},

where $g_{\mathrm{atom}}$, $g'_{\mathrm{atom}}$, $g_{\mathrm{bond}}$, and $g'_{\mathrm{bond}}$ are linear layers; $\sigma_r$ is the ReLU activation function. The $\parallel$ symbol indicates vector concatenation. To compute the final action probabilities, we apply the Softmax activation function to \emph{concatenated} vectors of logits of all possible atom actions $Act_a$ and possible bond actions $Act_b$.}

\paragraph{GCN-att layer}
\label{par:gcn_att}

The basic building block of the encoder and decoder is an attention-based GCN layer that we call \emph{GCN-att}. We enhance the GCN layer from \citet{gat} by adding bond features as input information for computing the attention values. Let $\mathbf{H}_{t} \in {\mathbb{R}^{n\times h}}$ denote input node features for the $t$-th GCN-att layer and $N(i) \subset \mathbb{Z}_{\geq 0}$ denote set of indices of neighbors of node at index $i$ (where $i \in N(i)$). We calculate new node features $\mathbf{H}_{t+1} \in {\mathbb{R}^{n\times h}}$ as follows:

\begin{align}
    \mathbb{R}^{d} \ni {\mathbf{H}^t_i}' &= \sigma_r(f^{t}_{att}(\mathbf{H}^t_i)) \\
    \mathbb{R}^{2d+n_b} \ni \mathbf{B}^t_{ij} &= {\mathbf{H}^t_i}' \parallel {\mathbf{H}^t_j}' \parallel \mathbf{A}_{i, j} \\
    \mathbb{R}^{K} \ni \mathbf{C}^t_{ij} &= f^{t}_{att'}(\mathbf{B}^t_{ij}) \\
    \mathbb{R}^{n_a} \ni \mathbf{G}^t_{ik} &= \underset{j \in N(i)}\sum \frac{\exp{C^t_{ijk}}}{\underset{l \in N(i)}\sum \exp{C^t_{ilk}}} {\mathbf{H}^t_j} \\
    \mathbb{R}^{n_a} \ni \mathbf{H}^{t+1}_i &= \underset{1 \leq k \leq K}\parallel \sigma_r(f^t_k(\mathbf{G}^t_i))
\end{align}

where as before $\sigma_r$ denotes ReLU activation function and $\parallel$ indicates vector concatenation. The scalar $K \in \mathbb{N}_{+}$ is the number of attention heads and $f_{att} \colon \mathbb{R}^{n_a} \to \mathbb{R}^{d}$, $f_{att'} \colon \mathbb{R}^{2d+n_b} \to \mathbb{R}^K$ and $f \colon \mathbb{R}^{n_a} \to \mathbb{R}^{n_a/K}$ are standard linear layers. Numbers $n_a$, $n_b$, $d$ and $K$ are hyperparameters of the model. We require that $n_a$ is divisible by $K$. We use the same hyperparameter values for all GCN-att layers in the model.

}

\subsection{Gradient-based training of MEGAN} \label{sec:gradientbasedtraining}

\sj{In contrast to \citet{kien2019},} who use reinforcement learning to train their model, we back-propagate directly through the maximum likelihood objective to train MEGAN. We use teacher forcing to train the model, i.e. to predict each step during reaction generation we use previous steps from the ground-truth as input to the model.

This is nontrivial, as computing the gradient of the likelihood objective requires defining a fixed ordering of actions~\citep{you2018}. To solve this issue, \citet{you2018} enumerates atoms using breadth-first search. We adapt a similar idea to reaction generation. 

At first, we define a general ordering of action types for forward- and retrosynthesis. This ordering defines which types of actions should be performed first if there are more than one type of actions that could be applied to an atom in a generation step. {There are various possible orderings, described in detail in the Section "Ablation study on action ordering". The highest accuracy is achieved in the ordering called \textsc{BFS Rand-AT}. In this ordering, for} retrosynthesis, we give bond deletion the highest priority, as it is usually the step that determines the reaction center, which is the first step of reaction prediction in other methods~\cite{g2g, graph_retro, yan2020, wang2020}. Analogously, adding a bond has the highest priority for forward synthesis, as this usually determines the reaction center in forward prediction. Priorities of other types of actions were determined experimentally; we did not observe a significant change in validation error when modifying these priorities. We list these priorities explicitly in Table~\ref{tab:action_type_priorities}.

The remaining ties between actions are broken as follows. For retrosynthesis prediction, we prioritize actions that act on the least recently modified atoms. Our motivation is to prioritize more difficult actions (such as deciding where the reaction occurs) at the expense of simpler actions (such as completing the synthons resulting from breaking a bond). In forward prediction, we prioritize the most frequently visited atom, which we found to perform better empirically. The remaining ties are broken at random. We describe the algorithm in full detail in the Supplement~(\nameref{supp:action_ordering}).

\begin{table}
\caption{Action prioritization used for training MEGAN, for forward and retrosynthesis.}
\label{tab:action_type_priorities}
\begin{center}
\begin{small}
\begin{sc}
\begin{tabular}{r | l | l}
\toprule
\multicolumn{3}{c}{Priorities of action types} \\
\midrule
& Retrosynthesis & Forward synthesis \\
\midrule
1. & \emph{EditBond (delete)} & \emph{EditBond (add)} \\
2. & \emph{EditBond (add)} & \emph{EditBond (delete)}  \\
3. & \emph{EditBond (other)} & \emph{EditBond (other)}  \\
4. & \emph{EditAtom} & \emph{EditAtom} \\
5. & \emph{AddBenzene} & \emph{AddBenzene} \\
6. & \emph{AddAtom} & \emph{AddAtom} \\

\bottomrule
\end{tabular}
\end{sc}
\end{small}
\end{center}
\vskip -0.2in
\end{table}

It is worth noting that the action ordering algorithm is used only to acquire ground truth samples for training MEGAN with a gradient-based method. The model is expected to learn a strong prior towards outputting edits in such order; however, when evaluating MEGAN, we do not explicitly limit the available action space, giving it a chance to determine the order by itself. We observe that MEGAN often predicts the ground truth reaction using a different order of actions than the ground truth training sample.

\sj{We use the following rules to resolve conflicts when the action could be outputted at different atoms or bonds:

\begin{itemize}
\item \emph{Stop} action can be predicted only by the supernode
\item Bond actions can be predicted for indices $i$ and $j$, where $i<j$ and nodes at $i$ and $j$ are atoms
\end{itemize}

Finally, for simplicity, we do not mask out redundant actions, such as deleting a non-existing bond or editing the atom to the same values of properties, as we expect the model to learn not to use such actions.}

\section{Experiments}
\label{experiments}

We evaluate MEGAN on three standard datasets for retro and forward synthesis prediction. We first evaluate it on the standard retrosynthesis prediction benchmark USPTO-50k. Next, we investigate how MEGAN scales to the large-scale retrosynthesis task. Finally, we present results on forward synthesis prediction on the USPTO-MIT dataset.

\subsection{Retrosynthesis}

\paragraph{Data} First, we evaluate MEGAN performance in retrosynthesis prediction. The goal is to predict correctly the set of reactants based on the product of a reaction. The accuracy is measured by comparing the SMILES~\citep{smiles} representation of the predicted molecules to the SMILES representation of the ground truth target molecules. Before the comparison, we remove atom mapping information from the SMILES strings and canonicalize them using RdKit~\citep{rdkit}. We use top-k accuracy computed on reactions from the test set as the main evaluation metric.

\ms{We evaluate on the USPTO-50k dataset of approximately 50000 reactions, which was collected by \citet{lowe} and classified into 10 reaction types by \citet{schneider}. We use the same processed version of the dataset as \citet{gln}, where each reaction consists of a single product molecule and a set of one or more reactants, with corresponding atoms between the reactants and the product mapped. Following other studies, we split the reactions into training/validation/test sets with respective sizes of 80\%/10\%/10\%. We use the dataset and split provided by \citet{gln}.}

\ms{
\begin{table}[H]
\caption{Top-k test accuracy for retrosynthesis prediction on the USPTO-50k dataset. MEGAN achieves state-of-the-art results for K equal, or larger than 1 compared to prior work, which demonstrates an excellent coverage of the chemical reaction space. Results of other methods are taken from \citet{gln,graph_retro,yan2020,wang2020, at2020}. $\dagger$ denotes concurrent work with entries separated by the horizontal line. \sj{We also note that AT averages predictions over 100 different augmentations, which significantly increases inference time.} The best results are bolded.}
\label{uspto_50k_topk}
\begin{center}
\begin{small}
\begin{sc}
\begin{tabular}{c c c c c c c}
\toprule
\multirow{2}{*}{Methods} & \multicolumn{6}{c}{Top-k accuracy \%} \\
 & 1 & 3 & 5 & 10 & 20 & 50 \\
\midrule
\multicolumn{7}{c}{Reaction type unknown} \\
\midrule
Trans & 37.9 & 57.3 & 62.7 & / & / & / \\
Retrosim & 37.3 & 54.7 & 63.3 & 74.1 & 82.0 & 85.3 \\
Neuralsym & 44.4 & 65.3 & 72.4 & 78.9 & 82.2 & 83.1 \\
GLN & \textbf{52.5} & 69.0 & 75.6 & 83.7 & 89.0 & 92.4 \\
MEGAN & 48.1 & \textbf{70.7} & \textbf{78.4} & \textbf{86.1} & \textbf{90.3} & \textbf{93.2} \\
\midrule
G2Gs $\dagger$ & 48.9 & 67.6 & 72.5 & 75.5 & / & / \\
RetroPrime $\dagger$ & 51.4 & \textbf{70.8} & 74.0 & 76.1 & / & / \\
AT (100x) $\dagger$ & \textbf{53.2} & / & \textbf{80.5} & 85.2 & / & / \\

\midrule
\multicolumn{7}{c}{Reaction type given as prior} \\
\midrule
Seq2seq & 37.4 & 52.4 & 57.0 & 61.7 & 65.9 & 70.7 \\
Retrosim & 52.9 & 73.8 & 81.2 & 88.1 & 91.8 & 92.9 \\
Neuralsym & 55.3 & 76.0 & 81.4 & 85.1 & 86.5 & 86.9 \\
GLN & \textbf{64.2} & 79.1 & 85.2 & 90.0 & 92.3 & 93.2 \\
MEGAN & 60.7 & \textbf{82.0} & \textbf{87.5} & \textbf{91.6} & \textbf{93.9} & \textbf{95.3} \\
\midrule
G2Gs $\dagger$ & 61.0 & 81.3 & 86.0 & 88.7 & / & / \\
RetroPrime $\dagger$ & \textbf{64.8} & 81.6 & 85.0 & 86.9 & / & / \\
\bottomrule
\end{tabular}
\end{sc}
\end{small}
\end{center}
\vskip -0.2in
\end{table}
}

\paragraph{Experimental setting} We run two variants of training on USPTO-50k: one with unknown reaction type and one for which reaction type is given as a prior by an additional embedding layer. For both runs, we use the same model architecture and the same training setup. The training takes approximately 16 hours on a single Nvidia Tesla K80 GPU for both variants.

We use beam search~\citep{beam_search} on output probabilities of actions to generate multiple ranked candidates for each product. For USPTO-50k, we set the maximum number of steps to 16 and the beam width to 50, as it is the largest $K$ for which accuracy was reported for the baseline models. 

For training, we use a batch size of 4 reactions. We use Adam optimizer~\citep{kingma2014method} with the initial learning rate of 0.0001. We use warm-up, increasing the learning rate from 0 to 0.0001 over the first 20000 training steps. For efficiency, we evaluate the validation loss on a subset of 2500 validation samples after each 20000 training samples. We multiply the learning rate by 0.1 if the estimated validation loss has not decreased for the last 4 evaluations. We stop the training after the estimated validation loss has not decreased in the last 8 evaluations. We adapt the hyperparameters based on the validation loss. The final hyperparameter values are described in the Supplement~(\nameref{supp:hyperparameter_search}).

\paragraph{Baselines} We compare the performance of MEGAN on USPTO-50k with several template-free and template-based models, including current state-of-the-art methods. Seq2seq~\citep{seq2seq} and Transformer~\citep{mt_retro} are both template-free methods based on machine translation models applied on SMILES strings.  Retrosim~\citep{retrosim} uses reaction fingerprint to select a template based on similar reactions in the dataset. Neuralsym~\citep{neuralsym} uses a multi-linear perceptron to rank templates. GLN~\citep{gln} employs a graph model that assesses when rules from templates should be applied.

We also include a comparison to concurrently developed methods. G2G~\citep{g2g} is a template-free model that, similarly to MEGAN, generates reactions by modifying molecular graphs but with a separate module for predicting reaction center. RetroPrime~\cite{wang2020} uses Transformer-based models for both the reaction center and leaving group prediction.  \ms{Augmented Transformer (AT) uses the Transformer architecture for template-free reaction generation with some additional SMILES augmentation techniques that improve test Top K accuracy but incurs additional inference cost due to the added test-time augmentation~\citep{at2020}.} \ms{We excluded from the comparison some of the concurrent works which had an information leak problem in evaluation (see: Section~\nameref{supp:excluded_concurrent} in the Supplement).}

\paragraph{Results} Table~\ref{uspto_50k_topk} reports results on the USPTO-50k benchmark in a variant with and without reaction type information provided. When the reaction type is given, MEGAN outperforms all baselines in all metrics, with the exception of top-1. \ms{When the reaction type is unknown, MEGAN beats prior models for $K \geq 3$ (and concurrent models for $K \geq 10$) and achieves comparable results for lower values of $K$.}

We hypothesize that the advantage of MEGAN for large $K$ stems largely from the fact that MEGAN generates reaction as a sequence of edits. This might help to efficiently search through different plausible reaction centers, hence covering a more diverse subset of the reaction space. It can also enable MEGAN to achieve high coverage of the reaction space, which is indicated by the top-50 accuracy of 93.2\% when reaction type is unknown and 95.3\% when the reaction type is provided.

\subsection{Large scale retrosynthesis prediction task}

\paragraph{Data} The USPTO-50k benchmark, although of relatively high quality, is a small dataset containing only 10 specific types of reactions. We train MEGAN for retrosynthesis on a large benchmark to test its scalability. We use the original data set collected by~\citet{lowe} containing reactions from US patents dating from 1976 to September 2016. We use the same preprocessed and split data as~\cite{gln}, which consists of approximately 800k/100k/100k training/validation/test reactions. We refer to this dataset as USPTO-FULL.

\paragraph{Experimental Setting} We train MEGAN on USPTO-FULL using the same architecture and training procedure as for USPTO-50k. We only increase the maximum number of actions from 16 to 32 to account for more complex reactions in the dataset. We use a beam search with beam width of 50 for evaluation. Otherwise, we use the same hyperparameters as on USPTO-50k. The training took about 60 hours on a single Tesla K80 GPU. 

\paragraph{Results} Table~\ref{uspto_full_retro} shows top-k accuracy on USPTO-FULL for MEGAN compared to other methods for retrosynthesis prediction. We see that our model achieves competitive performance on a large scale retrosynthesis data set, slightly outperforming other methods in terms of top-10 accuracy.

\begin{table}
\caption{Top-k test accuracy for retrosynthesis prediction on the USPTO-FULL dataset. Results for other methods are taken from~\citet{gln}~and~\citet{at2020}. $\dagger$ denotes concurrent work with entries separated by the horizontal line. \sj{We also note that AT averages predictions over 100 different augmentations, which significantly increases inference time.} The best results are bolded.}
\label{uspto_full_retro}
\begin{center}
\begin{small}
\begin{sc}
\begin{tabular}{c c c c}
\toprule
\multirow{2}{*}{Methods} & \multicolumn{3}{c}{Top-k accuracy \%} \\
 & 1 & 10 & 50 \\
\midrule
Retrosim & 32.8 & 56.1 & / \\
Neuralsym & 35.8 & 60.8 & / \\
GLN & 39.3 & 63.7 & / \\
MEGAN & 33.6 & \textbf{63.9} & 74.1 \\
\midrule
\ms{AT (100x)} $\dagger$ & \textbf{44.4} & \textbf{70.4} & \ \\
\bottomrule
\end{tabular}
\end{sc}
\end{small}
\end{center}
\vskip -0.2in
\end{table}

\subsection{Forward synthesis}
\paragraph{Data} Finally, we evaluate MEGAN on forward synthesis. The goal is to predict correctly the target based on the substrates. We train the model on a standard forward synthesis benchmark of approximately 480000 atom-mapped chemical reactions, split into training/validation/test sets of 410k/30k/40k samples, which we call USPTO-MIT~\citep{weisfeiler}.

\paragraph{Baselines} We compare MEGAN to following baselines. S2S~\citep{schwaller_seq2seq} and MT~\citep{mt} use machine translation models to predict the SMILES of the product from the SMILES of the substrates. WLDN~\citep{weisfeiler} identifies pairwise atom interactions in the reaction center and ranks enumerated feasible bond configurations between these atoms. WLDN5~\citep{coley2018reactivity} improves this method by combining the problems of reaction center prediction and candidate ranking into a single task. GTPN~\citep{kien2019} predicts actions on the graph of substrates, similarly to MEGAN, however, it limits them to actions between existing atom pairs. \sj{We include also the concurrent AT model~\citep{at2020} in the comparison}.

\paragraph{Experimental Setting} 

We remove the \textsc{Chiral tag} and \textsc{Bond stereo} features from the model input, as USPTO-MIT has no stereochemical information. Forward synthesis usually takes fewer modifications to predict than retrosynthesis, so we reduce the maximum number of steps to 8 and use a beam size of 20 for evaluation. We also change the strategy to break ties in action ordering. For forward synthesis, we prioritize actions that act on most recently modified atoms. Apart from these changes, we use the same architecture, hyperparameters, and training procedure as for USPTO-50k. 

Similarly to \citet{mt}, we train for two variants of forward synthesis prediction. For the \emph{separated} variant, compounds that directly contribute to the product are explicitly marked in the set of substrates with an additional atom feature. For the \emph{mixed} variant, such information is not provided, so the model has a harder task as it has to determine the reaction center from a larger number of possible reactants. 

\ms{In forward synthesis, MEGAN can generate multiple products for a given set of input substrates set, whereas the USPTO-MIT set contains single-product reactions. We use a simple heuristic to acquire the main reaction product from the MEGAN prediction by selecting the product that has the longest SMILES string. From our observation, this heuristic works correctly for most of the reactions and does not lower the Top K accuracies.}

\paragraph{Results} Table~\ref{uspto_mit_top1} shows how MEGAN compares with other methods on both variants in terms of top-1 accuracy. In contrast to retrosynthesis, MEGAN is slightly outperformed by Molecular Transformer~\citep{mt} on both forward synthesis tasks with the test accuracy of 89.3\% for \emph{separated} and 86.3\% for \emph{mixed}. This might be attributable to the larger action space (the number of atoms and bond choices in each step) in MEGAN in the forward direction compared to the backward direction, which might hinder training. The action space is especially large on the \emph{mixed} variant in which the substrates can include molecules with atoms that are not present in the target. Developing a better training strategy for forward prediction is a promising topic for future work.

In Table~\ref{uspto_mit_topk} we compare the performance of MEGAN and Molecular Transformer on USPTO-MIT in top $K$ predictions. We use the best non-ensemble models provided by~\citet{mt} and set the beam size to 20 during evaluation. We observe that MEGAN surpasses the accuracy of Molecular Transformer for high $K$ values, which again indicates its ability to explore the reaction space efficiently.

\begin{table}
\caption{Top-1 test accuracy for forward synthesis on the USPTO-MIT dataset. Results for other methods taken from \citet{mt} and \citet{at2020}. $\dagger$ denotes concurrent work.}
\label{uspto_mit_top1}
\begin{center}
\begin{small}
\begin{sc}
\begin{tabular}{l c c c c c c c}
\toprule
Variants & S2S & WLDN & GTPN & WLDN5 & MT & \ms{AT} $\dagger$ & MEGAN  \\
\midrule
Separated & 80.3 & 79.6 & 82.4 & 85.6 & \textbf{90.4} & \textbf{91.9} & 89.3 \\
Mixed &  & 74.0 & & & \textbf{88.6} & \textbf{90.4} & 86.3 \\
\bottomrule
\end{tabular}
\end{sc}
\end{small}
\end{center}
\vskip -0.2in
\end{table}

\begin{table}
\caption{Top-k test accuracy of MEGAN and Molecular Transformer for forward synthesis on the USPTO-MIT dataset.}
\label{uspto_mit_topk}
\begin{center}
\begin{small}
\begin{sc}
\begin{tabular}{c c c c c c c}
\toprule
\multirow{1}{*}{Methods} & \multicolumn{5}{c}{Top-k accuracy \%} \\
 & 1 & 2 & 3 & 5 & 10 & 20 \\
\midrule
\multicolumn{7}{c}{Separated} \\
MT & \textbf{90.5} & \textbf{93.7} & \textbf{94.7} & 95.3 & 96.0 & 96.5 \\
MEGAN & 89.3 & 92.7 & 94.4 & \textbf{95.6} & \textbf{96.7} & \textbf{97.5} \\
\midrule
\multicolumn{7}{c}{Mixed} \\
MT & \textbf{88.7} &\textbf{92.1}& \textbf{93.1} & \textbf{94.2} & 94.9 & 95.4 \\
MEGAN & 86.3 & 90.3 & 92.4 & 94.0 & \textbf{95.4} & \textbf{96.6} \\
\bottomrule
\end{tabular}
\end{sc}
\end{small}
\end{center}
\vskip -0.1in
\end{table}

\section{Analysis and ablation studies}

\subsection{Error analysis}

To better understand model performance, we perform here an error analysis focusing on retrosynthesis prediction on the USPTO-50k dataset. 

\begin{figure}
\begin{center}
\caption{The top-ranked retrosynthesis prediction by MEGAN for 5 random target molecules from the USPTO-50k test set on which the prediction is different from the ground truth reaction. The first four reactions are feasible and can be executed using standard methods. The last reaction may pose difficulty because of the nontrivial regioselectivity that depends on the exact reaction conditions utilized.}
\label{fig:5random}
\includegraphics[width=0.95\textwidth]{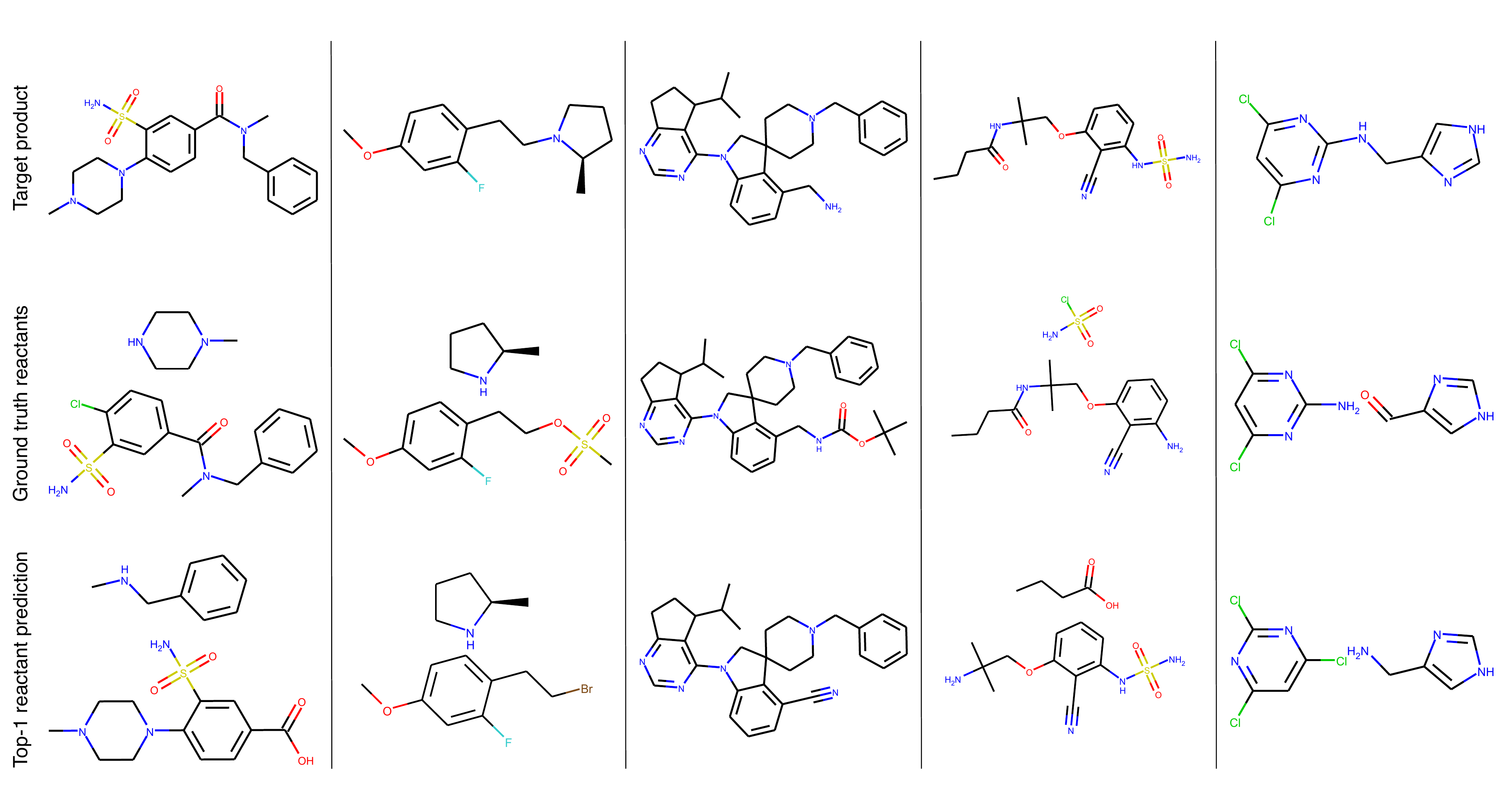}
\end{center}
\end{figure}

We first visualize in Figure~\ref{fig:5random} MEGAN prediction on the USPTO-50k test set for random 5 target molecules on which the predicted substrates are different from the ground truth reaction. We observe that the first four reactions are feasible and can be executed using standard methods. The last reaction may pose difficulty because of the nontrivial regioselectivity that depends on the exact reaction conditions utilized.

To better quantify the phenomenon that the top proposed reaction is often correct despite being different from the ground truth, we sampled random 100 reactions from the USPTO-50k test set for which the predicted substrates by MEGAN differed from the ground truth substrates. We then analyzed the correctness, by labeling them into the following categories: (1) no issues detected (seen as a true chemical reaction by a chemist), (2) incorrect chirality of the substrates, (3) the reaction has a low yield and/or important side products, (4) the reaction can be executed but only in multiple stages (e.g. with use of protecting groups), (5) there is a reactive functional group that participates in the reaction instead, (6) incorrect reaction for another reason. All 200 reactions were shuffled and assigned randomly to two experienced organic chemists.

\begin{figure}
\begin{center}
\includegraphics[width=0.8\textwidth]{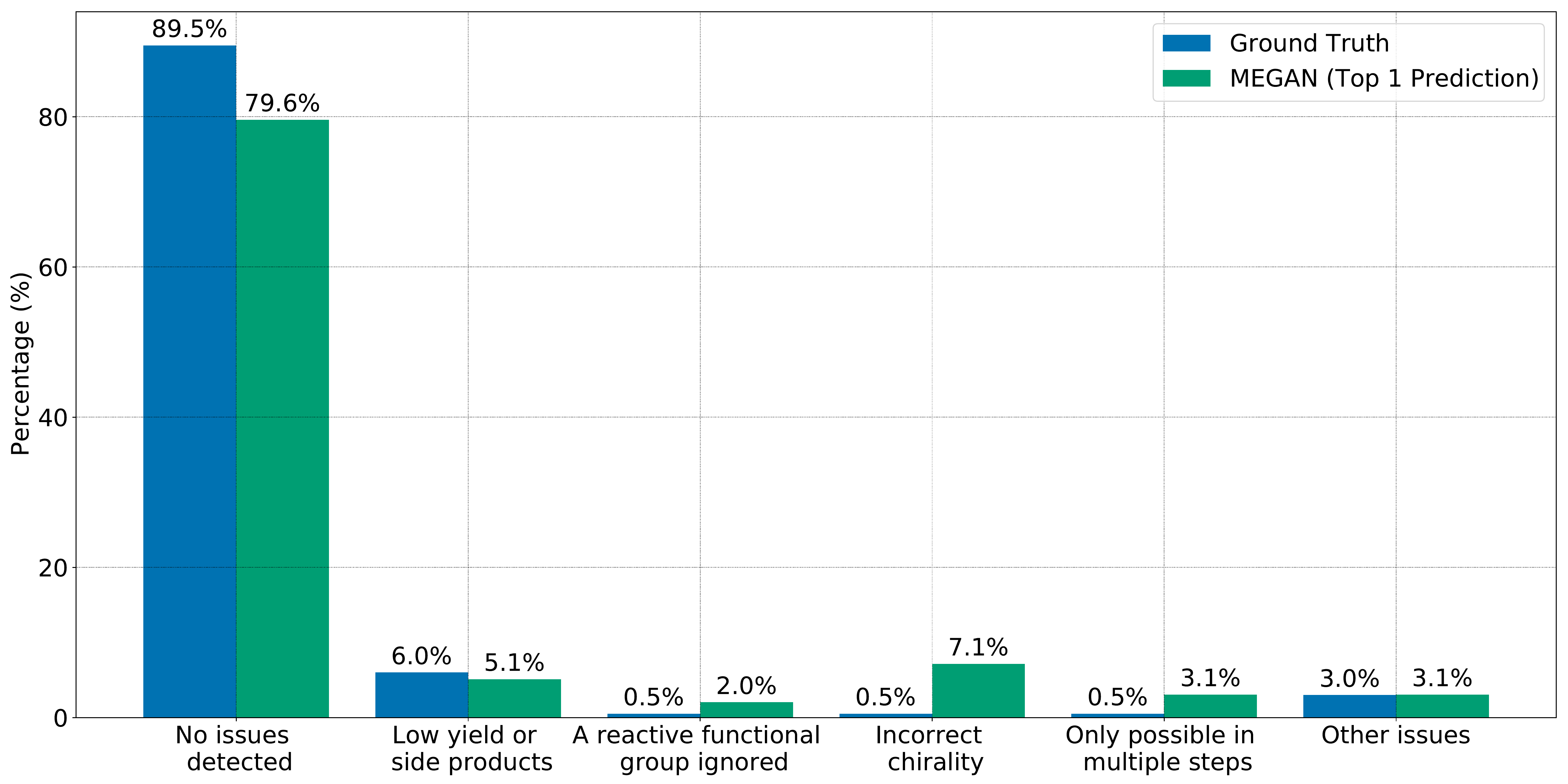}
\caption{Categorization of the top-ranked retrosynthesis predictions by MEGAN on a subset of USPTO-50k test set on which the top prediction differs from the ground truth reaction. For reference, we also include labels assigned to ground truth reaction from the test set. We find that the top-ranked prediction is often (in $79.6\%$ cases) deemed correct by a chemist even though it differs from the ground truth ("No issues detected"). The most common source of error is incorrectly predicting chirality in the substrates ("Incorrect chirality", $7.1\%$).}
\label{fig:erroranalysis}
\end{center}
\end{figure}

We show the break-down of MEGAN errors by these categories in Figure~\ref{fig:erroranalysis} and examples of reaction in each category in Figure~\ref{fig:erroranalysis}. On the whole, in $79.6\%$ cases, the top prediction by MEGAN was deemed correct by chemists even though it differed from the ground truth prediction. In comparison, $89.5\%$ of ground truth reactions were considered correct. Incorrect chirality of the substrates was the most common reason for reaction to be labeled as incorrect ($7.1\%$, compared to $0.5\%$ of ground truth reactions).

The main outcome of this error analysis is that for retrosynthesis prediction on the USPTO-50k, MEGAN performance seems to be close to human-level performance in the sense that the proposed reactions start to be hard to distinguish from ground-truth reactions in terms of their correctness. Having said that, there is still significant room for improvement, for example in terms of the chirality of the proposed substrates. Another common source of errors was ignoring the existence of a reactive functional group in the substrates.

\begin{figure}
\begin{center}
\caption{Examples of top-ranked retrosynthesis prediction by MEGAN for different error categories (c.f. Figure~\ref{fig:erroranalysis}).}
\label{fig:error_categories}
\includegraphics[width=0.9\textwidth]{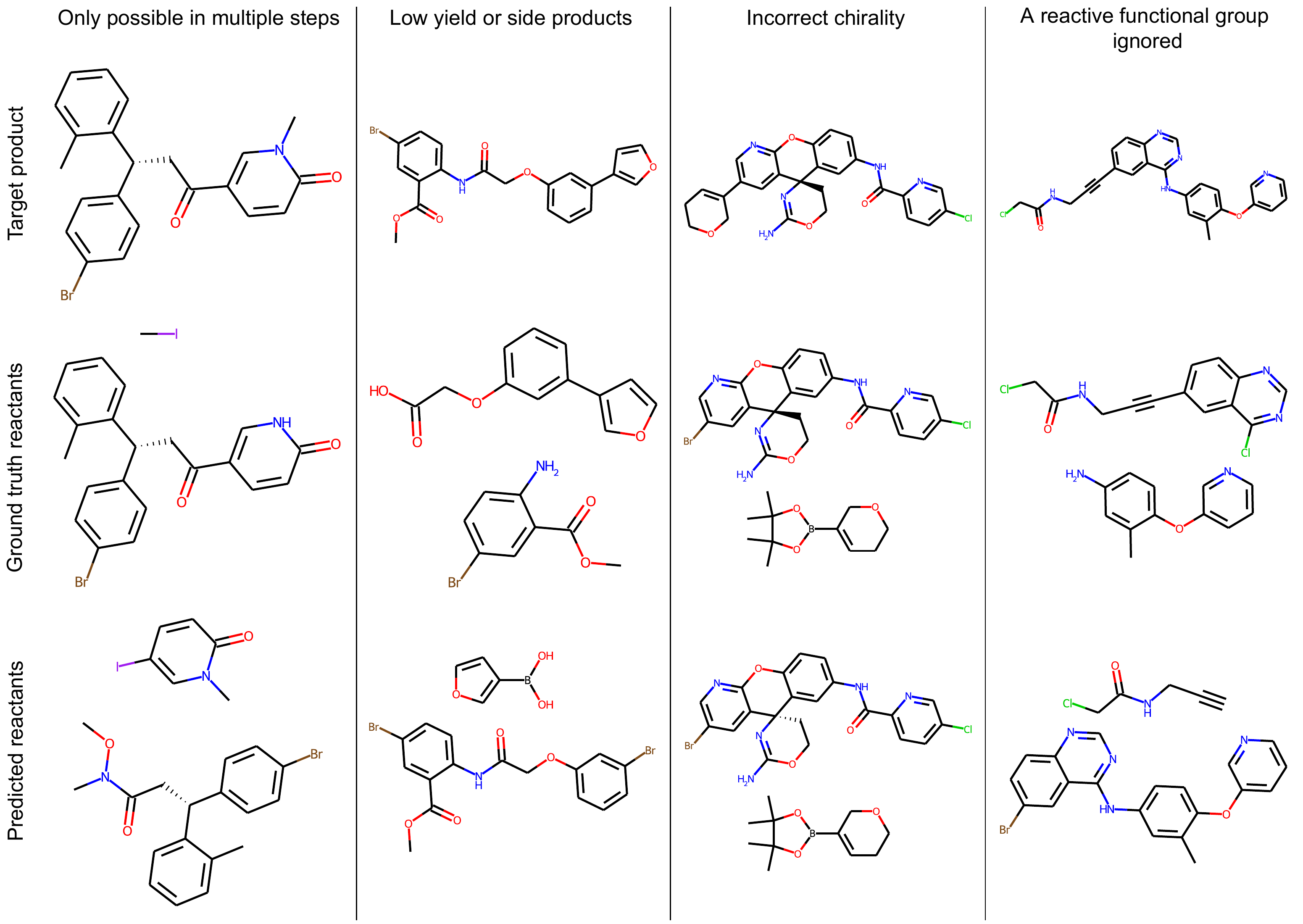}
\end{center}
\end{figure}

\subsection{Ablation study on action ordering}
\label{ablation_section}

Our gradient-based method of training MEGAN is based on a fixed ordering on ground-truth actions. Defining this ordering requires breaking ties between different actions of the same type. For example, if there are several \emph{EditBond} actions, we have to decide which should be executed first by the model.

In retrosynthesis prediction tasks, we broke ties based on the last time a given atom was modified. Actions that modify atoms that were least frequently modified are prioritized. The remaining ties were broken at random. We refer here to this ordering algorithm as \textsc{BFS Rand-AT}.

We explore here two modifications to this ordering. First, instead of prioritizing actions that modify the least frequently modified atom, we experiment with prioritizing actions that modify the most frequently modified atom. We refer to this as \text{DFS} ordering. Second, we investigate a strategy to break remaining ties based on the canonical SMILES. In this strategy, we prioritize actions that modify atoms that are earlier in the ordering determined by the canonical SMILES. We refer to this as \textsc{CANO-AT}. By combining these different choices, we arrive at four different ordering strategies: \textsc{BFS Rand-AT}, \textsc{DFS Rand-AT}, \textsc{BFS CANO-AT}, and \textsc{DFS CANO-AT}.

We present the performance of all variants on the USPTO-50k dataset in Table~\ref{ablation}. We observe that \textsc{BFS rand-at} ordering achieves the highest performance across most values of $K$, which motivates our choice to use it for training MEGAN on retrosynthesis prediction tasks. It is worth emphasizing that the choice of action ordering can strongly impact the performance of MEGAN. For example, the \textsc{random} action ordering achieves over $4\%$ lower top-1 accuracy compared to \textsc{DFS Rand-at}, which we used in the experiments.

\begin{table}
\caption{Top-k \ms{validation} accuracy on USPTO-50k (reaction type unknown) for different methods of ordering actions on the train set.}
\label{ablation}
\begin{center}
\begin{small}
\begin{sc}
\begin{tabular}{c c c c c c c}
\toprule
\multirow{2}{*}{Action order} & \multicolumn{6}{c}{Top-k accuracy \%} \\
 & 1 & 3 & 5 & 10 & 20 & 50 \\
\midrule
DFS cano-at & 47.6 & 71.2 & 79.5 & 87.0 & 91.4 & 94.2 \\
BFS cano-at & 47.5 & 71.8 & 80.2 & 87.0 & 91.4 & \textbf{94.5}  \\
DFS rand-at & 43.9 & 67.9 & 76.4 & 83.8 & 88.4 & 92.4 \\
BFS rand-at & \textbf{48.6} & \textbf{72.2} & \textbf{80.3} & \textbf{87.6} & \textbf{91.6} & 94.2 \\
random & 44.0 & 61.7 & 69.8 & 78.5 & 84.4 & 89.5  \\
\bottomrule
\end{tabular}
\end{sc}
\end{small}
\end{center}
\vskip -0.2in
\end{table}

\subsection{Performance with respect to reaction popularity}

Next, we analyze how MEGAN performance depends on reaction popularity. To approximate reaction popularity, we count the number of times that the corresponding template occurred in the training set. We use code provided by~\citet{coley2018reactivity} to extract templates for each ground-truth reaction from the datasets.

\begin{figure}[H]
\begin{center}
\includegraphics[width=0.8\textwidth]{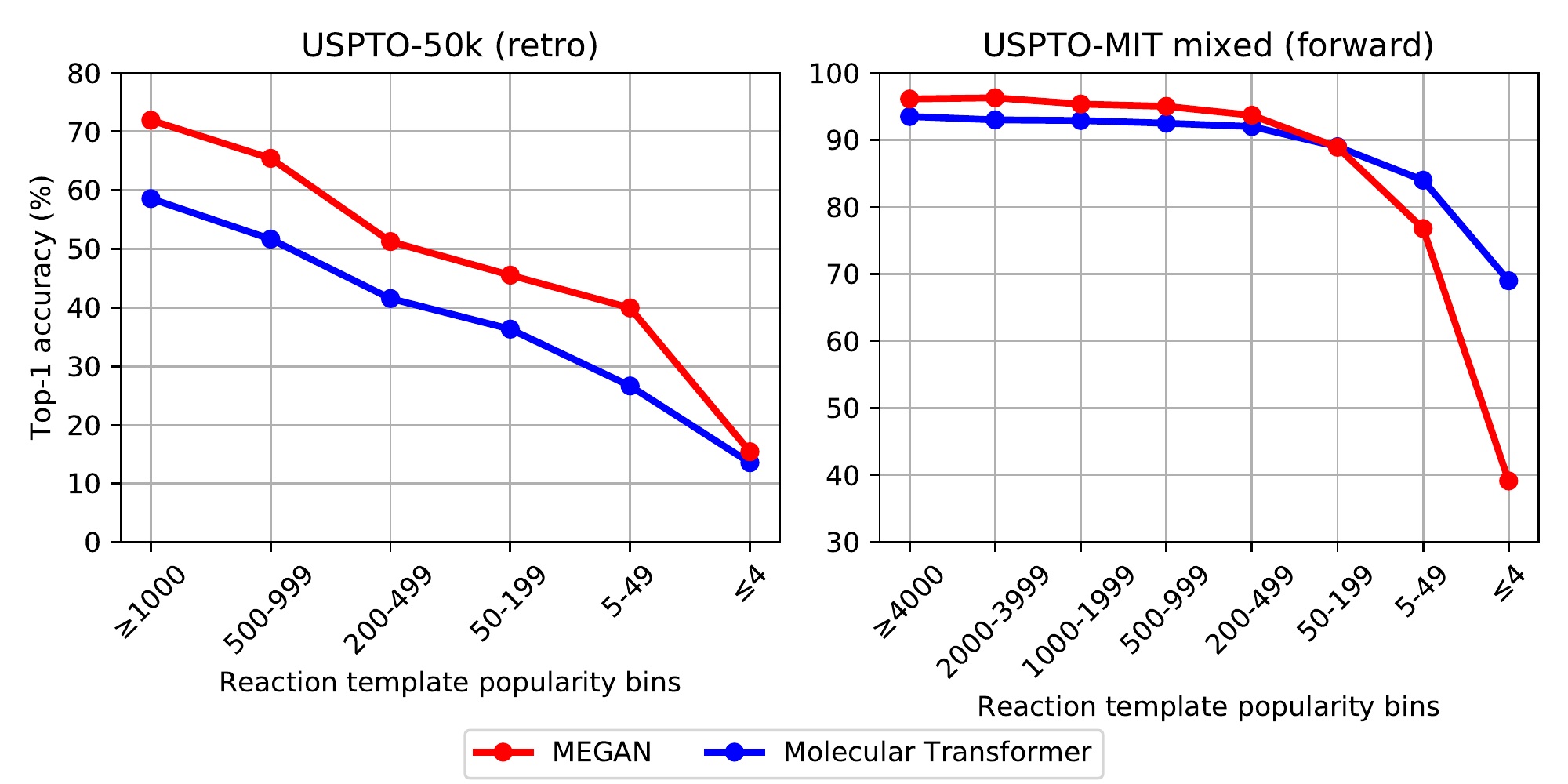}
\caption{Top-1 test accuracy of MEGAN and Molecular Transformer for reactions grouped by their corresponding reaction template popularity. MEGAN achieves higher accuracy on reactions that can be described by popular reaction templates on both USPTO-50k (left) and USPTO-MIT (right).}
\label{fig:temp_pop_acc}
\end{center}
\end{figure}

Figure~\ref{fig:temp_pop_acc} compares the test accuracy of MEGAN and Molecular Transformer depending on the ground-truth reaction type popularity. We use the model provided by~\citet{mt_retro} to get Transformer predictions on USPTO-50k and results from~\citet{mt} for comparison on USPTO-MIT. We observe that MEGAN performs better on popular types of reactions and underperforms compared to Molecular Transformer on the rarest reactions. A natural topic for future work is to improve the performance of MEGAN on this subset of the reaction space.

\section{Conclusions}
\label{conclusions}

In this work, we proposed Molecule Edit Graph Attention Network (MEGAN), a template-free model for both retrosynthesis and forward synthesis inspired by how chemists represent chemical reactions. MEGAN achieves competitive performance on both retrosynthesis and forward synthesis as well as state-of-the-art top-k accuracy for large K values on all tested datasets. MEGAN can be also scaled up to large reaction datasets.

Crucially, MEGAN generates reaction as a sequence of graph edits, which is inspired by how chemists represent chemical reactions. This inductive bias might explain the strong empirical performance of the model. In particular, we argued that it enables the model to more efficiently search through the space of plausible reactions. Looking forward, generating reactions as a sequence of edits is promising for building more intuitive human-computer interaction in synthesis planning software. 

A natural topic for the future is reducing the reliance of MEGAN on the mapping between products and substrates. We also found that incorrectly predicting chirality \sj{of the output} was the most common source of errors on the USPTO-50k dataset. Improving these aspects has the potential to further push state-of-the-art in retro and forward synthesis prediction.

\section{Data and Software Availability}

We open-sourced our code and trained models at \href{https://github.com/molecule-one/megan}{https://github.com/molecule-one/megan}. Our implementation is based on freely available software and is released under the MIT license. All datasets are available online, with references given in the main text.

\bibliography{megan}

\providecommand{\latin}[1]{#1}
\makeatletter
\providecommand{\doi}
  {\begingroup\let\do\@makeother\dospecials
  \catcode`\{=1 \catcode`\}=2 \doi@aux}
\providecommand{\doi@aux}[1]{\endgroup\texttt{#1}}
\makeatother
\providecommand*\mcitethebibliography{\thebibliography}
\csname @ifundefined\endcsname{endmcitethebibliography}
  {\let\endmcitethebibliography\endthebibliography}{}
\begin{mcitethebibliography}{55}
\providecommand*\natexlab[1]{#1}
\providecommand*\mciteSetBstSublistMode[1]{}
\providecommand*\mciteSetBstMaxWidthForm[2]{}
\providecommand*\mciteBstWouldAddEndPuncttrue
  {\def\EndOfBibitem{\unskip.}}
\providecommand*\mciteBstWouldAddEndPunctfalse
  {\let\EndOfBibitem\relax}
\providecommand*\mciteSetBstMidEndSepPunct[3]{}
\providecommand*\mciteSetBstSublistLabelBeginEnd[3]{}
\providecommand*\EndOfBibitem{}
\mciteSetBstSublistMode{f}
\mciteSetBstMaxWidthForm{subitem}{(\alph{mcitesubitemcount})}
\mciteSetBstSublistLabelBeginEnd
  {\mcitemaxwidthsubitemform\space}
  {\relax}
  {\relax}

\bibitem[Blakemore \latin{et~al.}(2018)Blakemore, Castro, Churcher, Rees,
  Thomas, Wilson, and Wood]{blake2018}
Blakemore,~D.~C.; Castro,~L.; Churcher,~I.; Rees,~D.~C.; Thomas,~A.~W.;
  Wilson,~D.~M.; Wood,~A. Organic synthesis provides opportunities to transform
  drug discovery. \emph{Nature Chemistry} \textbf{2018}, \emph{10},
  383--394\relax
\mciteBstWouldAddEndPuncttrue
\mciteSetBstMidEndSepPunct{\mcitedefaultmidpunct}
{\mcitedefaultendpunct}{\mcitedefaultseppunct}\relax
\EndOfBibitem
\bibitem[Corey and Wipke(1969)Corey, and Wipke]{corey}
Corey,~E.~J.; Wipke,~W.~T. Computer-Assisted Design of Complex Organic
  Syntheses. \emph{Science} \textbf{1969}, \emph{166}, 178--192\relax
\mciteBstWouldAddEndPuncttrue
\mciteSetBstMidEndSepPunct{\mcitedefaultmidpunct}
{\mcitedefaultendpunct}{\mcitedefaultseppunct}\relax
\EndOfBibitem
\bibitem[Segler \latin{et~al.}(2018)Segler, Preuss, and Waller]{segler2018}
Segler,~M. H.~S.; Preuss,~M.; Waller,~M.~P. Planning chemical syntheses with
  deep neural networks and symbolic AI. \emph{Nature} \textbf{2018},
  \emph{555}, 604--610\relax
\mciteBstWouldAddEndPuncttrue
\mciteSetBstMidEndSepPunct{\mcitedefaultmidpunct}
{\mcitedefaultendpunct}{\mcitedefaultseppunct}\relax
\EndOfBibitem
\bibitem[Coley \latin{et~al.}(2018)Coley, Green, and Jensen]{coley_ml_in_synth}
Coley,~C.~W.; Green,~W.~H.; Jensen,~K.~F. Machine Learning in Computer-Aided
  Synthesis Planning. \emph{Accounts of Chemical Research} \textbf{2018},
  \emph{51}, 1281--1289, PMID: 29715002\relax
\mciteBstWouldAddEndPuncttrue
\mciteSetBstMidEndSepPunct{\mcitedefaultmidpunct}
{\mcitedefaultendpunct}{\mcitedefaultseppunct}\relax
\EndOfBibitem
\bibitem[Lee \latin{et~al.}(2019)Lee, Yang, Sresht, Bolgar, Hou, Klug-McLeod,
  and Butler]{mt_unifies}
Lee,~A.~A.; Yang,~Q.; Sresht,~V.; Bolgar,~P.; Hou,~X.; Klug-McLeod,~J.~L.;
  Butler,~C.~R. Molecular Transformer unifies reaction prediction and
  retrosynthesis across pharma chemical space. \emph{Chem. Commun.}
  \textbf{2019}, \emph{55}, 12152--12155\relax
\mciteBstWouldAddEndPuncttrue
\mciteSetBstMidEndSepPunct{\mcitedefaultmidpunct}
{\mcitedefaultendpunct}{\mcitedefaultseppunct}\relax
\EndOfBibitem
\bibitem[Gao and Coley(2020)Gao, and Coley]{coley2020}
Gao,~W.; Coley,~C.~W. The Synthesizability of Molecules Proposed by Generative
  Models. \emph{Journal of Chemical Information and Modeling} \textbf{2020},
  \emph{60}, 5714--5723\relax
\mciteBstWouldAddEndPuncttrue
\mciteSetBstMidEndSepPunct{\mcitedefaultmidpunct}
{\mcitedefaultendpunct}{\mcitedefaultseppunct}\relax
\EndOfBibitem
\bibitem[Warren(2007)]{warren2007organic}
Warren,~S. \emph{Organic synthesis: the disconnection approach}; John Wiley \&
  Sons, 2007\relax
\mciteBstWouldAddEndPuncttrue
\mciteSetBstMidEndSepPunct{\mcitedefaultmidpunct}
{\mcitedefaultendpunct}{\mcitedefaultseppunct}\relax
\EndOfBibitem
\bibitem[Struble \latin{et~al.}(2020)Struble, Alvarez, Brown, Chytil, Cisar,
  DesJarlais, Engkvist, Frank, Greve, Griffin, Hou, Johannes, Kreatsoulas,
  Lahue, Mathea, Mogk, Nicolaou, Palmer, Price, Robinson, Salentin, Xing,
  Jaakkola, Green, Barzilay, Coley, and Jensen]{struble2020}
Struble,~T.~J. \latin{et~al.}  Current and Future Roles of Artificial
  Intelligence in Medicinal Chemistry Synthesis. \emph{Journal of Medicinal
  Chemistry} \textbf{2020}, \emph{63}, 8667--8682\relax
\mciteBstWouldAddEndPuncttrue
\mciteSetBstMidEndSepPunct{\mcitedefaultmidpunct}
{\mcitedefaultendpunct}{\mcitedefaultseppunct}\relax
\EndOfBibitem
\bibitem[Corey and Wipke(1969)Corey, and Wipke]{Corey178}
Corey,~E.~J.; Wipke,~W.~T. Computer-Assisted Design of Complex Organic
  Syntheses. \emph{Science} \textbf{1969}, \emph{166}, 178--192\relax
\mciteBstWouldAddEndPuncttrue
\mciteSetBstMidEndSepPunct{\mcitedefaultmidpunct}
{\mcitedefaultendpunct}{\mcitedefaultseppunct}\relax
\EndOfBibitem
\bibitem[Satoh and Funatsu(1995)Satoh, and Funatsu]{Satoh1995Sophia}
Satoh,~H.; Funatsu,~K. SOPHIA, a Knowledge Base-Guided Reaction Prediction
  System - Utilization of a Knowledge Base Derived from a Reaction Database.
  \emph{Journal of Chemical Information and Computer Sciences} \textbf{1995},
  \emph{35}, 34--44\relax
\mciteBstWouldAddEndPuncttrue
\mciteSetBstMidEndSepPunct{\mcitedefaultmidpunct}
{\mcitedefaultendpunct}{\mcitedefaultseppunct}\relax
\EndOfBibitem
\bibitem[Segler and Waller(2017)Segler, and Waller]{segler2017}
Segler,~M. H.~S.; Waller,~M.~P. Modelling Chemical Reasoning to Predict and
  Invent Reactions. \emph{Chemistry – A European Journal} \textbf{2017},
  \emph{23}, 6118--6128\relax
\mciteBstWouldAddEndPuncttrue
\mciteSetBstMidEndSepPunct{\mcitedefaultmidpunct}
{\mcitedefaultendpunct}{\mcitedefaultseppunct}\relax
\EndOfBibitem
\bibitem[Coley \latin{et~al.}(2017)Coley, Barzilay, Jaakkola, Green, and
  Jensen]{coley2017}
Coley,~C.~W.; Barzilay,~R.; Jaakkola,~T.~S.; Green,~W.~H.; Jensen,~K.~F.
  Prediction of Organic Reaction Outcomes Using Machine Learning. \emph{ACS
  Central Science} \textbf{2017}, \emph{3}, 434--443\relax
\mciteBstWouldAddEndPuncttrue
\mciteSetBstMidEndSepPunct{\mcitedefaultmidpunct}
{\mcitedefaultendpunct}{\mcitedefaultseppunct}\relax
\EndOfBibitem
\bibitem[Dai \latin{et~al.}(2019)Dai, Li, Coley, Dai, and Song]{gln}
Dai,~H.; Li,~C.; Coley,~C.; Dai,~B.; Song,~L. In \emph{Advances in Neural
  Information Processing Systems 32}; Wallach,~H., Larochelle,~H.,
  Beygelzimer,~A., d\textquotesingle Alch\'{e}-Buc,~F., Fox,~E., Garnett,~R.,
  Eds.; Curran Associates, Inc., 2019; pp 8872--8882\relax
\mciteBstWouldAddEndPuncttrue
\mciteSetBstMidEndSepPunct{\mcitedefaultmidpunct}
{\mcitedefaultendpunct}{\mcitedefaultseppunct}\relax
\EndOfBibitem
\bibitem[Grzybowski \latin{et~al.}(2018)Grzybowski, Szymku{\'c}, Gajewska,
  Molga, Dittwald, Wołoś, and Klucznik]{Grzybowski2018ChematicaAS}
Grzybowski,~B.; Szymku{\'c},~S.; Gajewska,~E.~P.; Molga,~K.; Dittwald,~P.;
  Wołoś,~A.; Klucznik,~T. Chematica: A Story of Computer Code That Started to
  Think like a Chemist. \emph{Chem} \textbf{2018}, \emph{4}, 390--398\relax
\mciteBstWouldAddEndPuncttrue
\mciteSetBstMidEndSepPunct{\mcitedefaultmidpunct}
{\mcitedefaultendpunct}{\mcitedefaultseppunct}\relax
\EndOfBibitem
\bibitem[Schwaller \latin{et~al.}(2019)Schwaller, Laino, Gaudin, Bolgar,
  Hunter, Bekas, and Lee]{mt}
Schwaller,~P.; Laino,~T.; Gaudin,~T.; Bolgar,~P.; Hunter,~C.~A.; Bekas,~C.;
  Lee,~A.~A. Molecular Transformer: A Model for Uncertainty-Calibrated Chemical
  Reaction Prediction. \emph{ACS Central Science} \textbf{2019}, \emph{5},
  1572–1583\relax
\mciteBstWouldAddEndPuncttrue
\mciteSetBstMidEndSepPunct{\mcitedefaultmidpunct}
{\mcitedefaultendpunct}{\mcitedefaultseppunct}\relax
\EndOfBibitem
\bibitem[Karpov \latin{et~al.}(2019)Karpov, Godin, and Tetko]{mt_retro}
Karpov,~P.; Godin,~G.; Tetko,~I. \emph{A Transformer Model for Retrosynthesis};
  2019; pp 817--830\relax
\mciteBstWouldAddEndPuncttrue
\mciteSetBstMidEndSepPunct{\mcitedefaultmidpunct}
{\mcitedefaultendpunct}{\mcitedefaultseppunct}\relax
\EndOfBibitem
\bibitem[Chen \latin{et~al.}(2019)Chen, Shen, Jaakkola, and
  Barzilay]{chen2019learning}
Chen,~B.; Shen,~T.; Jaakkola,~T.~S.; Barzilay,~R. Learning to Make
  Generalizable and Diverse Predictions for Retrosynthesis. \emph{arXiv
  e-prints} \textbf{2019}, arXiv--1910\relax
\mciteBstWouldAddEndPuncttrue
\mciteSetBstMidEndSepPunct{\mcitedefaultmidpunct}
{\mcitedefaultendpunct}{\mcitedefaultseppunct}\relax
\EndOfBibitem
\bibitem[Weininger(1988)]{smiles}
Weininger,~D. SMILES, a chemical language and information system. 1.
  Introduction to methodology and encoding rules. \emph{Journal of Chemical
  Information and Computer Sciences} \textbf{1988}, \emph{28}, 31--36\relax
\mciteBstWouldAddEndPuncttrue
\mciteSetBstMidEndSepPunct{\mcitedefaultmidpunct}
{\mcitedefaultendpunct}{\mcitedefaultseppunct}\relax
\EndOfBibitem
\bibitem[Roberts \latin{et~al.}(2020)Roberts, Liang, Neubig, and
  Lipton]{roberts2020decoding}
Roberts,~N.; Liang,~D.; Neubig,~G.; Lipton,~Z.~C. Decoding and Diversity in
  Machine Translation. 2020\relax
\mciteBstWouldAddEndPuncttrue
\mciteSetBstMidEndSepPunct{\mcitedefaultmidpunct}
{\mcitedefaultendpunct}{\mcitedefaultseppunct}\relax
\EndOfBibitem
\bibitem[He \latin{et~al.}(2018)He, Haffari, and
  Norouzi]{he-etal-2018-sequence}
He,~X.; Haffari,~G.; Norouzi,~M. Sequence to Sequence Mixture Model for Diverse
  Machine Translation. Proceedings of the 22nd Conference on Computational
  Natural Language Learning. Brussels, Belgium, 2018; pp 583--592\relax
\mciteBstWouldAddEndPuncttrue
\mciteSetBstMidEndSepPunct{\mcitedefaultmidpunct}
{\mcitedefaultendpunct}{\mcitedefaultseppunct}\relax
\EndOfBibitem
\bibitem[Jiang and de~Rijke(2018)Jiang, and
  de~Rijke]{jiang-de-rijke-2018-sequence}
Jiang,~S.; de~Rijke,~M. Why are Sequence-to-Sequence Models So Dull?
  Understanding the Low-Diversity Problem of Chatbots. Proceedings of the 2018
  {EMNLP} Workshop {SCAI}: The 2nd International Workshop on Search-Oriented
  Conversational {AI}. Brussels, Belgium, 2018; pp 81--86\relax
\mciteBstWouldAddEndPuncttrue
\mciteSetBstMidEndSepPunct{\mcitedefaultmidpunct}
{\mcitedefaultendpunct}{\mcitedefaultseppunct}\relax
\EndOfBibitem
\bibitem[Bradshaw \latin{et~al.}(2018)Bradshaw, Kusner, Paige, Segler, and
  Hernández-Lobato]{electron}
Bradshaw,~J.; Kusner,~M.~J.; Paige,~B.; Segler,~M. H.~S.;
  Hernández-Lobato,~J.~M. A Generative Model For Electron Paths. 2018;
  \url{https://arxiv.org/abs/1805.10970}\relax
\mciteBstWouldAddEndPuncttrue
\mciteSetBstMidEndSepPunct{\mcitedefaultmidpunct}
{\mcitedefaultendpunct}{\mcitedefaultseppunct}\relax
\EndOfBibitem
\bibitem[Do \latin{et~al.}(2019)Do, Tran, and Venkatesh]{kien2019}
Do,~K.; Tran,~T.; Venkatesh,~S. Graph Transformation Policy Network for
  Chemical Reaction Prediction. Proceedings of the 25th ACM SIGKDD
  International Conference on Knowledge Discovery \& Data Mining. New York, NY,
  USA, 2019; p 750–760\relax
\mciteBstWouldAddEndPuncttrue
\mciteSetBstMidEndSepPunct{\mcitedefaultmidpunct}
{\mcitedefaultendpunct}{\mcitedefaultseppunct}\relax
\EndOfBibitem
\bibitem[Gu \latin{et~al.}(2019)Gu, Liu, and Cho]{gui2019}
Gu,~J.; Liu,~Q.; Cho,~K. Insertion-based Decoding with Automatically Inferred
  Generation Order. \emph{Transactions of the Association for Computational
  Linguistics} \textbf{2019}, \emph{7}\relax
\mciteBstWouldAddEndPuncttrue
\mciteSetBstMidEndSepPunct{\mcitedefaultmidpunct}
{\mcitedefaultendpunct}{\mcitedefaultseppunct}\relax
\EndOfBibitem
\bibitem[Todd(2005)]{Todd2005}
Todd,~M.~H. Computer-aided organic synthesis. \emph{Chem. Soc. Rev.}
  \textbf{2005}, \emph{34}, 247--266\relax
\mciteBstWouldAddEndPuncttrue
\mciteSetBstMidEndSepPunct{\mcitedefaultmidpunct}
{\mcitedefaultendpunct}{\mcitedefaultseppunct}\relax
\EndOfBibitem
\bibitem[Salatin and Jorgensen(1980)Salatin, and Jorgensen]{salatin1980}
Salatin,~T.~D.; Jorgensen,~W.~L. Computer-assisted mechanistic evaluation of
  organic reactions. 1. Overview. \emph{The Journal of Organic Chemistry}
  \textbf{1980}, \emph{45}, 2043--2051\relax
\mciteBstWouldAddEndPuncttrue
\mciteSetBstMidEndSepPunct{\mcitedefaultmidpunct}
{\mcitedefaultendpunct}{\mcitedefaultseppunct}\relax
\EndOfBibitem
\bibitem[Zimmerman(2013)]{zimmerman2013}
Zimmerman,~P.~M. Automated discovery of chemically reasonable elementary
  reaction steps. \emph{Journal of Computational Chemistry} \textbf{2013},
  \emph{34}, 1385--1392\relax
\mciteBstWouldAddEndPuncttrue
\mciteSetBstMidEndSepPunct{\mcitedefaultmidpunct}
{\mcitedefaultendpunct}{\mcitedefaultseppunct}\relax
\EndOfBibitem
\bibitem[Molga \latin{et~al.}(2019)Molga, Gajewska, Szymkuć, and
  Grzybowski]{grzyb}
Molga,~K.; Gajewska,~E.~P.; Szymkuć,~S.; Grzybowski,~B.~A. The logic of
  translating chemical knowledge into machine-processable forms: a modern
  playground for physical-organic chemistry. \emph{React. Chem. Eng.}
  \textbf{2019}, \emph{4}, 1506--1521\relax
\mciteBstWouldAddEndPuncttrue
\mciteSetBstMidEndSepPunct{\mcitedefaultmidpunct}
{\mcitedefaultendpunct}{\mcitedefaultseppunct}\relax
\EndOfBibitem
\bibitem[Wei \latin{et~al.}(2016)Wei, Duvenaud, and Aspuru-Guzik]{guzik2016}
Wei,~J.~N.; Duvenaud,~D.; Aspuru-Guzik,~A. Neural Networks for the Prediction
  of Organic Chemistry Reactions. \emph{ACS Central Science} \textbf{2016},
  \emph{2}, 725--732\relax
\mciteBstWouldAddEndPuncttrue
\mciteSetBstMidEndSepPunct{\mcitedefaultmidpunct}
{\mcitedefaultendpunct}{\mcitedefaultseppunct}\relax
\EndOfBibitem
\bibitem[Jin \latin{et~al.}(2017)Jin, Coley, Barzilay, and
  Jaakkola]{weisfeiler}
Jin,~W.; Coley,~C.~W.; Barzilay,~R.; Jaakkola,~T. Predicting Organic Reaction
  Outcomes with Weisfeiler-Lehman Network. 2017;
  \url{https://arxiv.org/abs/1709.04555}\relax
\mciteBstWouldAddEndPuncttrue
\mciteSetBstMidEndSepPunct{\mcitedefaultmidpunct}
{\mcitedefaultendpunct}{\mcitedefaultseppunct}\relax
\EndOfBibitem
\bibitem[Coley \latin{et~al.}(2018)Coley, Jin, Rogers, Jamison, S~Jaakkola,
  Green, Barzilay, and Jensen]{coley2018reactivity}
Coley,~C.~W.; Jin,~W.; Rogers,~L.; Jamison,~T.~F.; S~Jaakkola,~T.;
  Green,~W.~H.; Barzilay,~R.; Jensen,~K.~F. A Graph-Convolutional Neural
  Network Model for the Prediction of Chemical Reactivity. 2018;
  \url{https://chemrxiv.org/articles/A_Graph-Convolutional_Neural_Network_Model_for_the_Prediction_of_Chemical_Reactivity/7163189/1}\relax
\mciteBstWouldAddEndPuncttrue
\mciteSetBstMidEndSepPunct{\mcitedefaultmidpunct}
{\mcitedefaultendpunct}{\mcitedefaultseppunct}\relax
\EndOfBibitem
\bibitem[Coley \latin{et~al.}(2017)Coley, Rogers, Green, and Jensen]{retrosim}
Coley,~C.~W.; Rogers,~L.; Green,~W.~H.; Jensen,~K.~F. Computer-Assisted
  Retrosynthesis Based on Molecular Similarity. \emph{ACS Central Science}
  \textbf{2017}, \emph{3}, 1237--1245, PMID: 29296663\relax
\mciteBstWouldAddEndPuncttrue
\mciteSetBstMidEndSepPunct{\mcitedefaultmidpunct}
{\mcitedefaultendpunct}{\mcitedefaultseppunct}\relax
\EndOfBibitem
\bibitem[Schwaller \latin{et~al.}(2018)Schwaller, Gaudin, Lányi, Bekas, and
  Laino]{schwaller_seq2seq}
Schwaller,~P.; Gaudin,~T.; Lányi,~D.; Bekas,~C.; Laino,~T. “Found in
  Translation”: predicting outcomes of complex organic chemistry reactions
  using neural sequence-to-sequence models. \emph{Chem. Sci.} \textbf{2018},
  \emph{9}, 6091--6098\relax
\mciteBstWouldAddEndPuncttrue
\mciteSetBstMidEndSepPunct{\mcitedefaultmidpunct}
{\mcitedefaultendpunct}{\mcitedefaultseppunct}\relax
\EndOfBibitem
\bibitem[Liu \latin{et~al.}(2017)Liu, Ramsundar, Kawthekar, Shi, Gomes, Nguyen,
  Ho, Sloane, Wender, and Pande]{seq2seq}
Liu,~B.; Ramsundar,~B.; Kawthekar,~P.; Shi,~J.; Gomes,~J.; Nguyen,~Q.~L.;
  Ho,~S.; Sloane,~J.; Wender,~P.; Pande,~V. Retrosynthetic reaction prediction
  using neural sequence-to-sequence models. \emph{ACS central science}
  \textbf{2017}, \emph{3}, 1103--1113\relax
\mciteBstWouldAddEndPuncttrue
\mciteSetBstMidEndSepPunct{\mcitedefaultmidpunct}
{\mcitedefaultendpunct}{\mcitedefaultseppunct}\relax
\EndOfBibitem
\bibitem[Mehri and Sigal(2018)Mehri, and Sigal]{Mehri2018MiddleOutD}
Mehri,~S.; Sigal,~L. Middle-Out Decoding. NeurIPS. 2018\relax
\mciteBstWouldAddEndPuncttrue
\mciteSetBstMidEndSepPunct{\mcitedefaultmidpunct}
{\mcitedefaultendpunct}{\mcitedefaultseppunct}\relax
\EndOfBibitem
\bibitem[Shi \latin{et~al.}(2020)Shi, Xu, Guo, Zhang, and Tang]{g2g}
Shi,~C.; Xu,~M.; Guo,~H.; Zhang,~M.; Tang,~J. A Graph to Graphs Framework for
  Retrosynthesis Prediction. 2020\relax
\mciteBstWouldAddEndPuncttrue
\mciteSetBstMidEndSepPunct{\mcitedefaultmidpunct}
{\mcitedefaultendpunct}{\mcitedefaultseppunct}\relax
\EndOfBibitem
\bibitem[Somnath \latin{et~al.}(2020)Somnath, Bunne, Coley, Krause, and
  Barzilay]{graph_retro}
Somnath,~V.~R.; Bunne,~C.; Coley,~C.~W.; Krause,~A.; Barzilay,~R. Learning
  Graph Models for Template-Free Retrosynthesis. 2020;
  \url{https://arxiv.org/abs/2006.07038}\relax
\mciteBstWouldAddEndPuncttrue
\mciteSetBstMidEndSepPunct{\mcitedefaultmidpunct}
{\mcitedefaultendpunct}{\mcitedefaultseppunct}\relax
\EndOfBibitem
\bibitem[Yan \latin{et~al.}(2020)Yan, Ding, Zhao, Zheng, Yang, Yu, and
  Huang]{yan2020}
Yan,~C.; Ding,~Q.; Zhao,~P.; Zheng,~S.; Yang,~J.; Yu,~Y.; Huang,~J. RetroXpert:
  Decompose Retrosynthesis Prediction like A Chemist. 2020;
  \url{https://chemrxiv.org/articles/preprint/Interpretable_Retrosynthesis_Prediction_in_Two_Steps/11869692/3}\relax
\mciteBstWouldAddEndPuncttrue
\mciteSetBstMidEndSepPunct{\mcitedefaultmidpunct}
{\mcitedefaultendpunct}{\mcitedefaultseppunct}\relax
\EndOfBibitem
\bibitem[Gao and Coley(2020)Gao, and Coley]{gao2020}
Gao,~W.; Coley,~C.~W. The Synthesizability of Molecules Proposed by Generative
  Models. 2020\relax
\mciteBstWouldAddEndPuncttrue
\mciteSetBstMidEndSepPunct{\mcitedefaultmidpunct}
{\mcitedefaultendpunct}{\mcitedefaultseppunct}\relax
\EndOfBibitem
\bibitem[Liu \latin{et~al.}(2020)Liu, Korablyov, Jastrz{\k{e}}bski,
  W{\l}odarczyk-Pruszy{\'n}ski, Bengio, and Segler]{liu2020retrognn}
Liu,~C.-H.; Korablyov,~M.; Jastrz{\k{e}}bski,~S.;
  W{\l}odarczyk-Pruszy{\'n}ski,~P.; Bengio,~Y.; Segler,~M.~H. RetroGNN:
  Approximating Retrosynthesis by Graph Neural Networks for De Novo Drug
  Design. \emph{arXiv preprint arXiv:2011.13042} \textbf{2020}, \relax
\mciteBstWouldAddEndPunctfalse
\mciteSetBstMidEndSepPunct{\mcitedefaultmidpunct}
{}{\mcitedefaultseppunct}\relax
\EndOfBibitem
\bibitem[Thakkar \latin{et~al.}(2021)Thakkar, Chadimová, Bjerrum, Engkvist,
  and Reymond]{Thakkar2021}
Thakkar,~A.; Chadimová,~V.; Bjerrum,~E.~J.; Engkvist,~O.; Reymond,~J.-L.
  Retrosynthetic accessibility score (RAscore) – rapid machine learned
  synthesizability classification from AI driven retrosynthetic planning.
  \emph{Chem. Sci.} \textbf{2021}, \emph{12}, 3339--3349\relax
\mciteBstWouldAddEndPuncttrue
\mciteSetBstMidEndSepPunct{\mcitedefaultmidpunct}
{\mcitedefaultendpunct}{\mcitedefaultseppunct}\relax
\EndOfBibitem
\bibitem[Gottipati \latin{et~al.}(2020)Gottipati, Sattarov, Niu, Pathak, Wei,
  Liu, Liu, Blackburn, Thomas, Coley, Tang, Chandar, and Bengio]{Gottipati2021}
Gottipati,~S.~K.; Sattarov,~B.; Niu,~S.; Pathak,~Y.; Wei,~H.; Liu,~S.; Liu,~S.;
  Blackburn,~S.; Thomas,~K.; Coley,~C.; Tang,~J.; Chandar,~S.; Bengio,~Y.
  Learning to Navigate The Synthetically Accessible Chemical Space Using
  Reinforcement Learning. Proceedings of the 37th International Conference on
  Machine Learning. 2020; pp 3668--3679\relax
\mciteBstWouldAddEndPuncttrue
\mciteSetBstMidEndSepPunct{\mcitedefaultmidpunct}
{\mcitedefaultendpunct}{\mcitedefaultseppunct}\relax
\EndOfBibitem
\bibitem[Veličković \latin{et~al.}(2017)Veličković, Cucurull, Casanova,
  Romero, Liò, and Bengio]{gat}
Veličković,~P.; Cucurull,~G.; Casanova,~A.; Romero,~A.; Liò,~P.; Bengio,~Y.
  Graph Attention Networks. 2017; \url{https://arxiv.org/abs/1710.10903}\relax
\mciteBstWouldAddEndPuncttrue
\mciteSetBstMidEndSepPunct{\mcitedefaultmidpunct}
{\mcitedefaultendpunct}{\mcitedefaultseppunct}\relax
\EndOfBibitem
\bibitem[Li \latin{et~al.}(2017)Li, Cai, and He]{supernode}
Li,~J.; Cai,~D.; He,~X. Learning Graph-Level Representation for Drug Discovery.
  2017; \url{https://arxiv.org/abs/1709.03741}\relax
\mciteBstWouldAddEndPuncttrue
\mciteSetBstMidEndSepPunct{\mcitedefaultmidpunct}
{\mcitedefaultendpunct}{\mcitedefaultseppunct}\relax
\EndOfBibitem
\bibitem[Landrum()]{rdkit}
Landrum,~G. RDKit: Open-source cheminformatics.
  \url{http://www.rdkit.org}\relax
\mciteBstWouldAddEndPuncttrue
\mciteSetBstMidEndSepPunct{\mcitedefaultmidpunct}
{\mcitedefaultendpunct}{\mcitedefaultseppunct}\relax
\EndOfBibitem
\bibitem[Vaswani \latin{et~al.}(2017)Vaswani, Shazeer, Parmar, Uszkoreit,
  Jones, Gomez, Kaiser, and Polosukhin]{attention}
Vaswani,~A.; Shazeer,~N.; Parmar,~N.; Uszkoreit,~J.; Jones,~L.; Gomez,~A.~N.;
  Kaiser,~L.; Polosukhin,~I. Attention Is All You Need. 2017;
  \url{https://arxiv.org/abs/1706.03762}\relax
\mciteBstWouldAddEndPuncttrue
\mciteSetBstMidEndSepPunct{\mcitedefaultmidpunct}
{\mcitedefaultendpunct}{\mcitedefaultseppunct}\relax
\EndOfBibitem
\bibitem[You \latin{et~al.}(2018)You, Ying, Ren, Hamilton, and
  Leskovec]{you2018}
You,~J.; Ying,~R.; Ren,~X.; Hamilton,~W.; Leskovec,~J. {G}raph{RNN}: Generating
  Realistic Graphs with Deep Auto-regressive Models. Proceedings of the 35th
  International Conference on Machine Learning. Stockholmsmässan, Stockholm
  Sweden, 2018; pp 5708--5717\relax
\mciteBstWouldAddEndPuncttrue
\mciteSetBstMidEndSepPunct{\mcitedefaultmidpunct}
{\mcitedefaultendpunct}{\mcitedefaultseppunct}\relax
\EndOfBibitem
\bibitem[Wang \latin{et~al.}(2020)Wang, Qiu, Li, Chen, Liu, Liao, Hsieh, and
  Yao]{wang2020}
Wang,~X.; Qiu,~J.; Li,~Y.; Chen,~G.; Liu,~H.; Liao,~B.; Hsieh,~C.-Y.; Yao,~X.
  RetroPrime: A Chemistry-Inspired and Transformer-based Method for
  Retrosynthesis Predictions. 2020;
  \url{https://chemrxiv.org/articles/preprint/RetroPrime_A_Chemistry-Inspired_and_Transformer-based_Method_for_Retrosynthesis_Predictions/12971942/1}\relax
\mciteBstWouldAddEndPuncttrue
\mciteSetBstMidEndSepPunct{\mcitedefaultmidpunct}
{\mcitedefaultendpunct}{\mcitedefaultseppunct}\relax
\EndOfBibitem
\bibitem[Lowe(2012)]{lowe}
Lowe,~D. Extraction of chemical structures and reactions from the literature.
  Ph.D.\ thesis, University of Cambridge, 2012\relax
\mciteBstWouldAddEndPuncttrue
\mciteSetBstMidEndSepPunct{\mcitedefaultmidpunct}
{\mcitedefaultendpunct}{\mcitedefaultseppunct}\relax
\EndOfBibitem
\bibitem[Schneider \latin{et~al.}(2016)Schneider, Stiefl, and
  Landrum]{schneider}
Schneider,~N.; Stiefl,~N.; Landrum,~G.~A. What’s What: The (Nearly)
  Definitive Guide to Reaction Role Assignment. \emph{Journal of Chemical
  Information and Modeling} \textbf{2016}, \emph{56}, 2336--2346, PMID:
  28024398\relax
\mciteBstWouldAddEndPuncttrue
\mciteSetBstMidEndSepPunct{\mcitedefaultmidpunct}
{\mcitedefaultendpunct}{\mcitedefaultseppunct}\relax
\EndOfBibitem
\bibitem[Tetko \latin{et~al.}(2020)Tetko, Karpov, Van~Deursen, and
  Godin]{at2020}
Tetko,~I.~V.; Karpov,~P.; Van~Deursen,~R.; Godin,~G. State-of-the-art augmented
  NLP transformer models for direct and single-step retrosynthesis.
  \emph{Nature Communications} \textbf{2020}, \emph{11}, 2041--1723\relax
\mciteBstWouldAddEndPuncttrue
\mciteSetBstMidEndSepPunct{\mcitedefaultmidpunct}
{\mcitedefaultendpunct}{\mcitedefaultseppunct}\relax
\EndOfBibitem
\bibitem[Graves(2012)]{beam_search}
Graves,~A. Sequence Transduction with Recurrent Neural Networks. 2012;
  \url{https://arxiv.org/abs/1211.3711}\relax
\mciteBstWouldAddEndPuncttrue
\mciteSetBstMidEndSepPunct{\mcitedefaultmidpunct}
{\mcitedefaultendpunct}{\mcitedefaultseppunct}\relax
\EndOfBibitem
\bibitem[Kingma and Ba(2014)Kingma, and Ba]{kingma2014method}
Kingma,~D.~P.; Ba,~J. Adam: A Method for Stochastic Optimization. 2014;
  \url{http://arxiv.org/abs/1412.6980}, cite arxiv:1412.6980Comment: Published
  as a conference paper at the 3rd International Conference for Learning
  Representations, San Diego, 2015\relax
\mciteBstWouldAddEndPuncttrue
\mciteSetBstMidEndSepPunct{\mcitedefaultmidpunct}
{\mcitedefaultendpunct}{\mcitedefaultseppunct}\relax
\EndOfBibitem
\bibitem[Segler and Waller(2017)Segler, and Waller]{neuralsym}
Segler,~M. H.~S.; Waller,~M.~P. Neural-Symbolic Machine Learning for
  Retrosynthesis and Reaction Prediction. \emph{Chemistry – A European
  Journal} \textbf{2017}, \emph{23}, 5966--5971\relax
\mciteBstWouldAddEndPuncttrue
\mciteSetBstMidEndSepPunct{\mcitedefaultmidpunct}
{\mcitedefaultendpunct}{\mcitedefaultseppunct}\relax
\EndOfBibitem
\end{mcitethebibliography}

\clearpage

\section{Supplementary Material}
\label{supplement}

\subsection{Featurization}
\label{supp:featurization}

\ms{We featurize atoms and bonds with one-hot encoded vectors of features calculated using RdKit~\citep{rdkit}. On each dataset, we select features that allow for correct reconstruction of all products and substrates from the development (training+validation) set. In Tables \ref{tab:atom_bond_features} we present all used atom and bond features and their possible values found on USPTO-50k. 

We concatenate one-hot feature vectors to gain the final input representation of atoms and bonds $\mathbf{H}^{OH} \in {\mathbb{Z}_{\geq 0}}^{n\times h_{OH}}$ and $\mathbf{A}^{OH} \in {\mathbb{Z}_{\geq 0}}^{n\times n\times a_{OH}}$. During the evaluation, if an unknown feature value is seen (for instance, \textsc{Bond~type=Quadruple}, which does not appear in the development set), we set the one-hot vector for this feature to zeros. For supernodes and all bonds connected with supernodes, all one-hot feature vectors are set to zeros, apart from vectors for the features \textsc{Is~supernode} and \textsc{Bond~type}. We connect each atom with itself with a special bond of type \textsc{Self}. For non-neighboring atoms at $i$ and $j$ we set $A^{OH}_{ij}=\vec{0}$}

\begin{table}
\caption{Features of atoms and bonds, \sj{with values they take on the USPTO-50k development set.}}
\label{tab:atom_bond_features}
\begin{minipage}{.5\linewidth}
\centering
\begin{small}
\begin{sc}
\begin{tabular}{p{2cm} p{4cm} p{0.5cm}}
\toprule
Name & Values \sj{on USPTO-50k} & Dim \\
\midrule
Is supernode & Yes, No & 2 \\
Atomic number & 5, 6, 7, 8, 9, 12, 14, 15, 16, 17, 29, 30, 34, 35, 50, 53 & 16 \\
Formal charge & -1, 0, 1 & 3 \\
Chiral tag & None, @, @@ & 3 \\
Number of explicit Hs & 0, 1, 2, 4 & 4 \\
Is aromatic & Yes, No & 2 \\
Is edited & Yes, No & 2 \\
\midrule
\multicolumn{2}{l}{Total atom feature vector length} & 32 \\
\bottomrule
\end{tabular}
\end{sc}
\end{small}
\end{minipage}\hfill
\begin{minipage}{.5\linewidth}
\centering
\begin{small}
\begin{sc}
\begin{tabular}{p{2cm} p{4cm} p{0.5cm}}
\toprule
Name & Possible values & Dim \\
\midrule
Bond type & Supernode, Self, Single, Double, Triple, Aromatic & 6 \\
Bond stereo & None, Z, E & 3 \\
Is edited & Yes, No & 2\\
\midrule
\multicolumn{2}{l}{Total bond feature vector length} & 11 \\
\bottomrule
\end{tabular}
\end{sc}
\end{small}
\end{minipage}
\end{table}

For both atoms and bonds, we add a special \textsc{Is~edited} feature that marks all bonds and atoms that have been modified by actions. This aims to help the decoder to focus on these atoms and bonds, as they are the most probable candidates for the next generation steps.

\subsection{Graph edit actions}
\label{supp:graph_edit_actions}

In Table~\ref{tab:all_actions} we show all possible graph actions found on the USPTO-50k development set. The actions were found during the generation of the training samples, which we describe in the next section. The actions have different sets of parameters, depending on the action type.

\begin{table*}
\caption{Graph actions found on the USPTO-50k development set.}
\label{tab:all_actions}
\begin{center}
\begin{small}
\begin{sc}
\begin{tabular}{c p{8cm} c}
\toprule
Action type & Action parameters & \# actions \\
\midrule
\emph{EditAtom} & Formal charge, Chiral tag, Num of explicit Hs,  Is aromatic & 11\\ 
\emph{EditBond} & Bond type, Bond stereo & 7 \\
\emph{AddAtom} & Atomic num, Formal charge, Chiral tag, Num of exp Hs, Is aromatic, Bond type, Bond stereo & 34 \\
\emph{AddBenzene} & / & 1 \\
\emph{Stop} & / & 1 \\
\midrule
 & & 54 \\
 \bottomrule
\end{tabular}
\end{sc}
\end{small}
\end{center}
\end{table*}

For \emph{EditAtom}, the action parameters are the atom properties that are changed by the action. Note that a single \emph{EditAtom} action sets all these properties to the specified values. For instance, action number 1, when executed on an atom, sets its formal charge to 0, chiral tag to None, number of explicit Hydrogen atoms to 1 and marks it as aromatic. \emph{EditBond} acts similarly to \emph{EditAtom} but edits properties of a bond instead of an atom. \emph{AddAtom} adds a new atom with specified features, connected to an existing atom with a bond with specified features. \emph{AddBenzene} appends a benzene ring to a specified carbon atom and has no parameters. \emph{Stop} terminates reaction generation and also has no parameters.

\emph{BondEdit} actions are bond actions, that is they are predicted for a pair of atoms. All other types of actions are atom actions and are predicted for a single atom in the graph.

\subsection{Ground truth action ordering}
\label{supp:action_ordering}

An important design choice in training of MEGAN is the algorithm that determines ordering of ground-truth actions that transform the product into the substrates (or vice versa in forward prediction). This ordering allows us to backpropagate through the maximum likelihood objective on the training set.

\paragraph{Action type priorities} At first, we define a general ordering of action types for forward- and retrosynthesis in Table~\ref{tab:action_type_priorities}. This ordering defines which types of actions should be performed first if there are more than one type of actions that could be applied to an atom in a generation step. For retrosynthesis, we give bond deletion the highest priority, as it is usually the step that determines the reaction center, which is the first step of reaction prediction in other methods~\cite{g2g, graph_retro, yan2020, wang2020}. Analogously, adding a bond has the highest priority for forward synthesis, as this usually determines the reaction center in forward prediction. Priorities of other types of actions were determined experimentally; we did not observe a significant change in validation error when modifying these priorities.

\paragraph{Atom priorities} 
Given the ordering of action types described in Table~\ref{tab:action_type_priorities}, we still need to decide on which action to choose in a scenario when two of the ground-truth graph edits are possible for different atoms. For example, we might need to decide on the order of adding leaving groups to two atoms that were disconnected by deleting a bond between them. Ordering of such actions can be determined by two factors:

\begin{enumerate}
\item Whether an atom has already been modified (\textsc{BFS} vs \textsc{DFS})
\item Order of the atom in the canonical SMILES of the target compounds (\textsc{cano-at} vs \textsc{rand-at})
\end{enumerate}

In the case of \emph{EditBond} actions that act on pairs of atoms, their priority is determined based on the highest of the priorities of these atoms. We present the whole algorithm for determining the next action in the Algorithm~\ref{alg:training_reaction_generation}.

\begin{algorithm}[H]
    \caption{Generating ground-truth actions for training samples}
    \label{alg:training_reaction_generation}
\begin{algorithmic} 
    \STATE {\bfseries Input:} input graph $I$, target graph $T$, ordering type
    \STATE Let $M(I) = $ \{all atoms that have been modified or added by previous actions\}
    \STATE Let $A(I) = $ \{all atoms in $I$ that could be edited with actions that lead to $T$\}
    \item[]
    \IF{$|A(I)| = 0$}
        \RETURN \emph{Stop}
    \ENDIF
    \item[]
    \IF{ordering type = \textsc{BFS cano-at} or ordering type = \textsc{BFS rand-at} } 
        \STATE Let $A^{M}(I) = A(I) \cup M(I) $
            
        \IF{|$A^{M}(I)$| > 0}
            \STATE Let $A'(I) = A^{M}(I)$
        \ELSE
            \STATE Let $A'(I) = A(I)$
        \ENDIF
    
    \ELSIF{ordering type = \textsc{DFS cano-at} or ordering type = \textsc{DFS rand-at} } 
        \STATE Let $A'(I) = A(I) \setminus M(I) $
    \ENDIF
    \item[]
    \IF{ordering type = \textsc{RANDOM}}
        \STATE Let $a$ = a random atom from $A(I)$
    \ELSIF{ordering type = \textsc{BFS cano-at} or ordering type = \textsc{DFS cano-at} } 
        \STATE Let $a$ = first atom from $A'(I)$ according to the atom order in canonical SMILES of $T$
    \ELSIF{ordering type = \textsc{BFS rand-at} or ordering type = \textsc{DFS rand-at} } 
        \STATE Let $a$ = a random atom from $A'(I)$
    \ENDIF
    \item[]
    \STATE Let $Act(a)$ = {all actions that lead to $T$ that could be performed on $a$}
    \STATE Let $Act'(a)$ = {all actions from $Act(a)$ with the action type that has the highest priority}
    \STATE Let $ActionType$ = action type of actions from $Act'(a)$

    \IF{$ActionType$ = \emph{EditAtom}}
        \STATE By definition, there could only be a single \emph{EditAtom} action performed on an atom.
        \RETURN The single \emph{EditAtom} action from $Act'(a)$
    \ELSIF{$ActionType$ = \emph{EditBond}}
        \STATE Let $b$ = the second atom from actions in $Act'(a)$ ranked highest according to the atom ordering
        \RETURN The \emph{EditBond} action between $a$ and $b$
    \ELSIF{$ActionType$ = \emph{AddAtom}}
        \STATE Let $b$ = the second atom from actions in $Act'(a)$ ranked highest according to the atom ordering
        \RETURN The \emph{AddAtom} action between $a$ and $b$
    \ELSIF{$ActionType$ = \emph{AddBenzene}}
        \RETURN A random \emph{AddBenzene} action from $Act'(a)$ 
    \ENDIF
\end{algorithmic}
\end{algorithm}

\subsection{Hyperparameter search}
\label{supp:hyperparameter_search}

Table~\ref{hyperparams} shows the hyperparameter values used for the final model on the USPTO-50k. We found these hyperparameters manually based on the performance on the USPTO-50k validation set. For other datasets, we only slightly modified these hyperparameters (as described in the main text).

\begin{table}[H]
\caption{Hyperparameters of the final model for the USPTO-50k dataset.}
\label{hyperparams}
\begin{center}
\begin{small}
\begin{sc}
\begin{tabular}{c | c | c | c | c | c | c}
\toprule
$n_h$ & $n_a$ & $n_d$ & $K$ & $n_e$ & $n_d$ & total params\\
1024 & 128 & 128 & 8 & 6 & 2 & \texttildelow 9.9 m \\
\bottomrule
\end{tabular}
\end{sc}
\end{small}
\end{center}
\end{table}

\ms{\subsection{Concurrent methods excluded from the comparisons}
\label{supp:excluded_concurrent}
After a detailed examination, we decided to exclude from the benchmark comparisons some of the concurrent methods of retrosynthesis prediction that suffer from an information leak problem between the training and test sets. This problem is described in details at \href{https://github.com/uta-smile/RetroXpert}{https://github.com/uta-smile/RetroXpert} (accessed on May 7th, 2021).
The methods with the information leak problem include GraphRetro~\citep{graph_retro} and RetroXpert~\cite{yan2020}. }

\end{document}